\documentclass[runningheads]{llncs}
\usepackage{graphicx}
\usepackage{tikz}
\usepackage{comment}
\usepackage{amsmath,amssymb} 
\usepackage{color}
\usepackage{multirow}
\usepackage{booktabs}
\usepackage{array}

\newcommand{\figfref}[1]{Figure~\ref{#1}}
\newcommand{\figref}[1]{Figure~\ref{#1}}

\newcommand{\tabref}[1]{Table~\ref{#1}}
\newcommand{\tabfref}[1]{Table~\ref{#1}}

\newcommand{\eqnref}[1]{(\ref{#1})}

\newcommand{\etal}{\textit{et al}.}
\newcommand{\ie}{\textit{i}.\textit{e}.}

\newcommand\blfootnote[1]{%
  \begingroup
  \renewcommand\thefootnote{}\footnote{#1}%
  \addtocounter{footnote}{-1}%
  \endgroup
}

\begin{document}
\title{Lightweight Alpha Matting Network Using Distillation-Based Channel Pruning}
\titlerunning{Distillation-Based Channel Pruning}
%
\author{Donggeun Yoon\inst{1}\orcidID{0000-0001-6153-8056} \and
Jinsun Park\inst{2}\orcidID{0000-0002-2296-819X} \and
Donghyeon Cho*\inst{1}\orcidID{0000-0002-2184-921X}}
\authorrunning{D. Yoon et al.}
%
\institute{Chungnam National University, Daejeon, South Korea \\
\email{202250187@o.cnu.ac.kr, cdh12242@cnu.ac.kr} \and
Pusan National University, Pusan, South Korea\\
\email{jspark@pnu.ac.kr}}
\maketitle              
\begin{abstract}
Recently, alpha matting has received a lot of attention because of its usefulness in mobile applications such as selfies.
%
%
Therefore, there has been a demand for a lightweight alpha matting model due to the limited computational resources of commercial portable devices.
%
%
To this end, we suggest a distillation-based channel pruning method for the alpha matting networks.
%
%
In the pruning step, we remove channels of a student network having fewer impacts on mimicking the knowledge of a teacher network.
%
%
Then, the pruned lightweight student network is trained by the same distillation loss.
%
%
A lightweight alpha matting model from the proposed method outperforms existing lightweight methods.
%
%
To show superiority of our algorithm, we provide various quantitative and qualitative experiments with in-depth analyses.
%
%
Furthermore, we demonstrate the versatility of the proposed distillation-based channel pruning method by applying it to semantic segmentation.

\end{abstract}
\blfootnote{ * Corresponding author.  \\ Project page is at \textcolor{magenta}{\tt \url{https://github.com/DongGeun-Yoon/DCP}}}
\begin{figure*}[t]
    \centering
    \begin{tabular}{@{}c@{\hskip 0.02\linewidth}c}
    \includegraphics[width=0.47\linewidth]{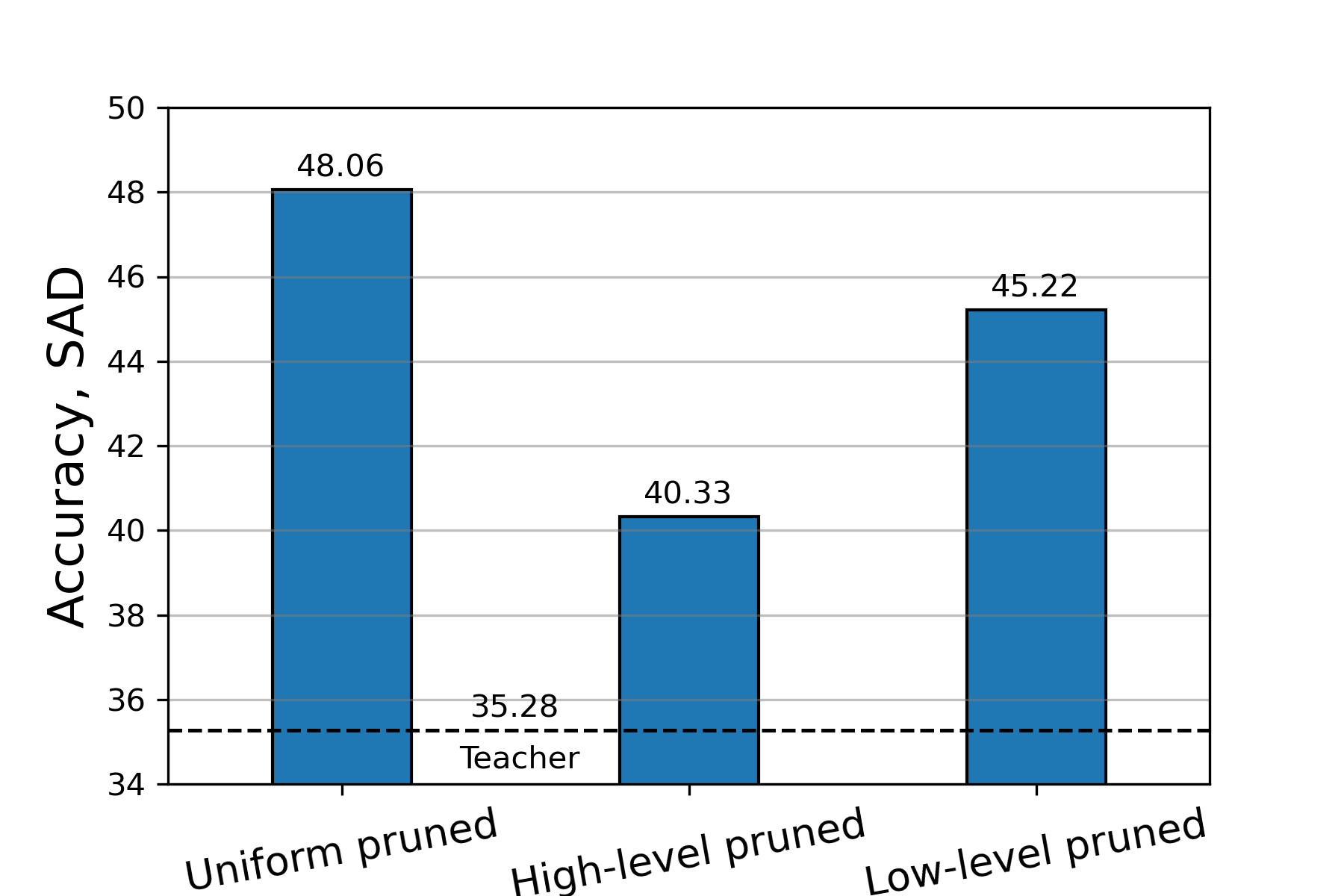} &
    \includegraphics[width=0.47\linewidth]{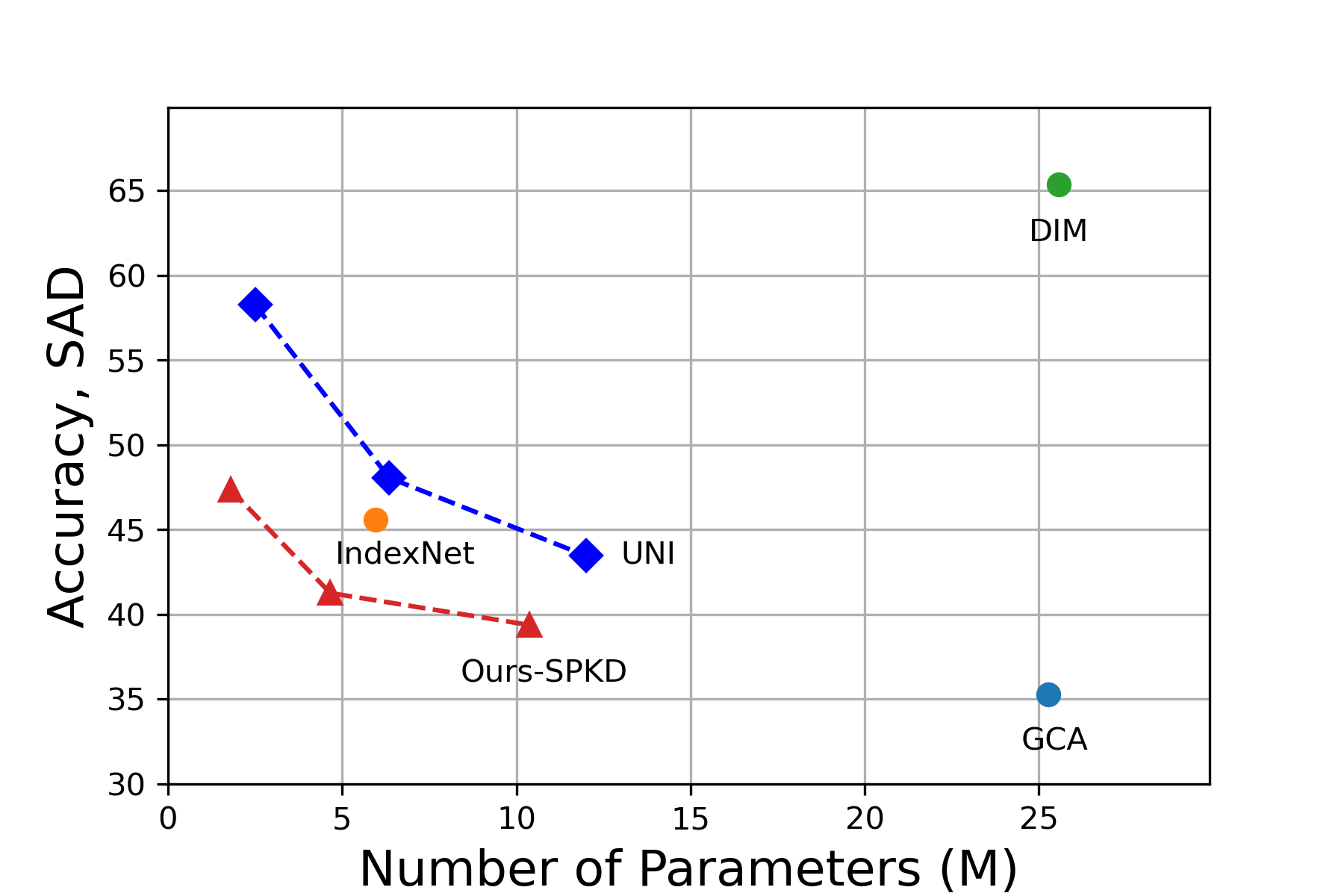}
    \end{tabular}
    \caption{
    (Left) Comparison of arbitrarily pruned models. The high-level pruned model removes channels from the high-level layers, and the low-level pruned model removes the channels from the low-level layers.  (Right) Trade-offs between accuracy and model size. Circle dots show results of original models. Squares and triangles denote results of pruned GCA models using uniform pruning and our method, respectively.
    }
    \label{fig:Teaser}
\end{figure*}
\section{Introduction}
The purpose of a natural image matting (\ie, alpha matting) is to estimate the transparency of the user-specified foreground in an image.
%
%
The alpha matting is formally defined as follows~\cite{CVPR2001Chuang}:
\begin{eqnarray}
I  =  \alpha  F + (1-\alpha)  B,
\label{eq:formulation}
\end{eqnarray}
where $I$, $F$ and $B$ are the observed color image, foreground, background, respectively. 
Also, $\alpha$ is transparency (\ie, alpha matte).
%
%
The natural image matting is a highly ill-posed problem because it needs to estimate $F$, $B$, and $\alpha$ simultaneously from an input color image $I$ and a trimap providing known foreground and background pixels.
Traditional approaches for natural image matting are categorized into affinity-based and sampling-based methods. 
Affinity-based methods~\cite{TOG05Sun,TPAMI08Levin,ICCV09Zheng,CVPR10He,CVPR11Lee,CVPR12Chen,Choi12eccv,CVPR13Chen,Aksoy17cvpr} propagate alpha values from known regions to pixels in unknown regions by analyzing statistical correlation among pixels.
Meanwhile, sampling-based methods~\cite{CVPR07Wang,ACMEG10Gastal,CVPR11He,CVPR2012Shahrian,CVPR13Shahrian,cho2014consistent,ICCV15Karacan,cho2016automatic} construct foreground and background color sample sets using pixels in known areas, then estimates alpha values in unknown regions.
%
%
However, these algorithms often rely on strong assumptions such as local smoothness~\cite{TPAMI08Levin} or sparsity of foreground and background colors~\cite{CVPR07Wang}.
Since the advent of large-scale image matting datasets such as Adobe-1k~\cite{Xu17cvpr}, deep learning-based matting algorithms been actively studied~\cite{Cho16eccv,Shen16eccv,cho2018deep,Hao19indexnet,Lutz18gan,Tang19learning_sampling,Cai19disentangled,Li20GCA,Yu20AGHSA,Sun_2021_CVPR}.
These methods outperform conventional ones remarkably.
Usually, the alpha matting networks are based on U-Net~\cite{UNET} or fully-convolutional networks (FCN)~\cite{FCN}.
For better performance, the number of layers or channels can be increased and also auxiliary modules can be added to baseline networks.
%
%
However, this leads to the increased computational costs and memory requirements that can be problematic for mobile applications.
Recently, a lightweight alpha matting network based on similarity-preserving knowledge distillation (SPKD)~\cite{yoon2020lightweight} was introduced to resolve these issues.
It successfully transfers similarities of features from the teacher network to the student network, which make the student network achieves much better performance than the baseline student network trained from scratch.
%
%

However, it is still an open problem that which architecture is the best one for the lightweight student network for natural image matting.
%
%
It can be seen from the left of~\figref{fig:Teaser} that the performance varies greatly depending on which layer the channels are removed from.
Note that the high-level pruned model has fewer parameters than the low-level pruned model.
%
%
Also, as shown in right of~\figref{fig:Teaser}, there is a trade-off between performance and model size, thus it is important to find an proper network architecture.
To find the optimal lightweight network architecture, various network pruning techniques can be applied.
Although it has been actively studied in the field of classification, it has not been dealt with much in the reconstruction problem including alpha matting.
Recently, channel pruning methods for semantic segmentation task were introduced in~\cite{Chen2020MultiTaskPF,He2021CAP}, but they mainly focus on preserving high-level semantic information rather than low-level fine structures that are crucial for the natural image matting problems.

%
%
%
%

To focus on low-level fine details during the channel pruning, we borrow the power of a pre-trained high-performance matting network which well preserves fine details.
In other words, in this paper, we present a distillation-based channel pruning method that removes the channels having a low impact in mimicking the pre-trained teacher network.
%
%
In the pruning phase, we induce the sparsity of the scaling factor of the batch normalization (BN) layer as in~\cite{Liu2017slimming,Chen2020MultiTaskPF,He2021CAP} and additionally apply distillation loss with a powerful pre-trained teacher model that is capable of precisely guiding a student network to preserve fine structural details in its prediction.
In the training phase, we train the pruned lightweight network by the same distillation loss used in the pruning stage.
Note that proposed method can make existing lightweight model (\ie, IndexNet) even lighter.

Our contributions can be summarized as follows.
(i) We introduce a novel channel pruning method for the natural image matting problem. To our best knowledge, this is the first attempt to apply the network pruning technique for the alpha matting problem.
(ii) By utilizing a distillation loss within the channel pruning step, we succeed in finding a lightweight alpha matting network that can recover fine details.
(iii) Our pruned network outperforms other baseline pruning approaches on two publicly available alpha matting datasets (Adobe-1K, Distinctions-646) while having a comparable number of parameters. In addition, we provide various ablation studies for a deeper understanding.
%
\section{Related Works}
\subsection{Natural Image Matting}
Most image matting techniques can be categorized into affinity-based and color sampling-based methods.
In affinity-based methods, statistical affinity is analyzed among the local and non-local neighbors to propagate values of alpha to the unknown areas from the known regions. 
%
%
Levin~\etal~\cite{TPAMI08Levin} introduced the closed-form solution based on matting Laplacian using the linear color model.
For handling high resolution images, He~\etal~\cite{CVPR10He} proposed efficient method to solve a large kernel matting Laplacian.
Furthermore, Lee and Wu~\cite{CVPR11Lee} introduced non-local matting propagating alpha values across non-local neighboring pixels.
Chen~\etal~\cite{CVPR12Chen} suggested the KNN matting which uses only $k$-nearest non-local neighbors to propagate alpha values. 
In addition, Chen~\etal~\cite{CVPR13Chen} utilized both local and non-local smoothness prior and Aksoy~\etal~\cite{Aksoy17cvpr} proposed multiple definitions of affinity for natural image matting.

The color sampling-based methods find foreground and background colors from constructed color sampler sets, then estimate alpha values in unknown regions.
Bayesian matting~\cite{CVPR2001Chuang} utilizes statistical models to analyze pixels in unknown regions.
Robust matting~\cite{CVPR07Wang}, shared matting~\cite{ACMEG10Gastal} and weighted color and texture matting~\cite{CVPR2012Shahrian} select the best color samples based on their own designed cost functions that take into account spatial, photometric, or texture information.
He~\etal~\cite{CVPR11He} proposed a randomized searching method to use global samples in the known areas to find the best combination of foreground and background colors.
Shahrian~\etal~\cite{CVPR13Shahrian} constructed comprehensive color sample sets to cover broad color variations using Gaussian Mixture Model (GMM).
Karacan~\etal~\cite{ICCV15Karacan} choose colors of foreground and background based on sparse representation.

After large-scale alpha matting datasets were published~\cite{Xu17cvpr,Yu20AGHSA}, a lot of deep learning-based works have been introduced.
Xu~\etal~\cite{Xu17cvpr} proposed a simple two-stage network for natural image matting.
Lutz~\etal~\cite{Lutz18gan} applied adversarial training for obtaining visually appealing alpha matte results.
To preserve details of alpha mattes, Hao~\etal~\cite{Hao19indexnet} introduced IndexNet including indices-guided unpooling operation. 
In addition, contextual attention~\cite{Li20GCA} and hierarchical attention~\cite{Yu20AGHSA} mechanisms were proposed for the matting problem.
Yu~\etal~\cite{Yu21MGM} proposed mask-guided matting leveraging a progressive refinement network with a general coarse mask as guidance.
Although the performance of alpha matting has been substantially improved, there are still not many studies on lightening alpha matting networks.
Recently, Yoon~\etal~\cite{yoon2020lightweight} succeeded to utilize knowledge distillation (KD) to obtain the lightweight deep learning model for alpha matting.
%
%
They reduce the number of channels with a fixed ratio, therefore, the optimal channel reduction ratio should be determined empirically.
%
%
\subsection{Network Pruning}
The purpose of the network pruning is to reduce redundancies in the over-parameterized deep convolutional neural network (CNN) models for fast run-time while maintaining performance.
In general, network pruning is divided into unstructured pruning~\cite{Han2015nips,Molchanov2017icml,frankle2018the,NEURIPS2020_46a4378f} which requires special libraries or hardware, and structured pruning~\cite{Wen2016SSL,Li2017pruning_filter,Changpinyo2017power_sparsity,Liu2017slimming} which is relatively easy to implement.
In this subsection, we focus on structured pruning that is more relevant to our work.
Wen~\etal~\cite{Wen2016SSL} proposed a Structured Sparsity Learning (SSL) method to sparsify structures including filters, channels, or layers by using group sparsity regularization.
Li~\etal~\cite{Li2017pruning_filter} introduced a method to remove channels having small incoming weights in a trained deep CNN model.
Changpinyo~\etal~\cite{Changpinyo2017power_sparsity} deactivate connections between filters in convolutional layers to obtain smaller networks.
Liu~\etal~\cite{Liu2017slimming} proposed the network slimming method to explicitly impose channel-wise sparsity in the deep CNN model using scaling factors in batch normalization.
%
Gao~\etal~\cite{gao2018dynamic} proposed a feature boosting and suppression (FBS) method to dynamically remove and boost channels according to the inputs using auxiliary connections.
%
Despite many pruning studies, most of them focus on the classification task. 
Fortunately, pruning techniques for semantic segmentation have begun to be introduced recently.
Chen~\etal~\cite{Chen2020MultiTaskPF} suggested a channel pruning method for semantic segmentation based on multi-task learning.
Furthermore, He~\etal~\cite{He2021CAP} proposed context-aware channel pruning method by leveraging the layer-wise channels interdependency.
However, pruning researches for matting network have not been addressed yet.

%
Since the estimation of the fine structures in alpha mattes are the most important objective of the matting network, a powerful pruning technique suitable for this purpose is strongly required.
%
To this end, we present a distillation-based channel pruning technique that exploits a powerful pre-trained alpha matting model suitable for recovering low-level fine details.

%
\begin{figure*}[t]
    \centering
    \begin{tabular}{@{}c}
    \includegraphics[width=0.94\linewidth]{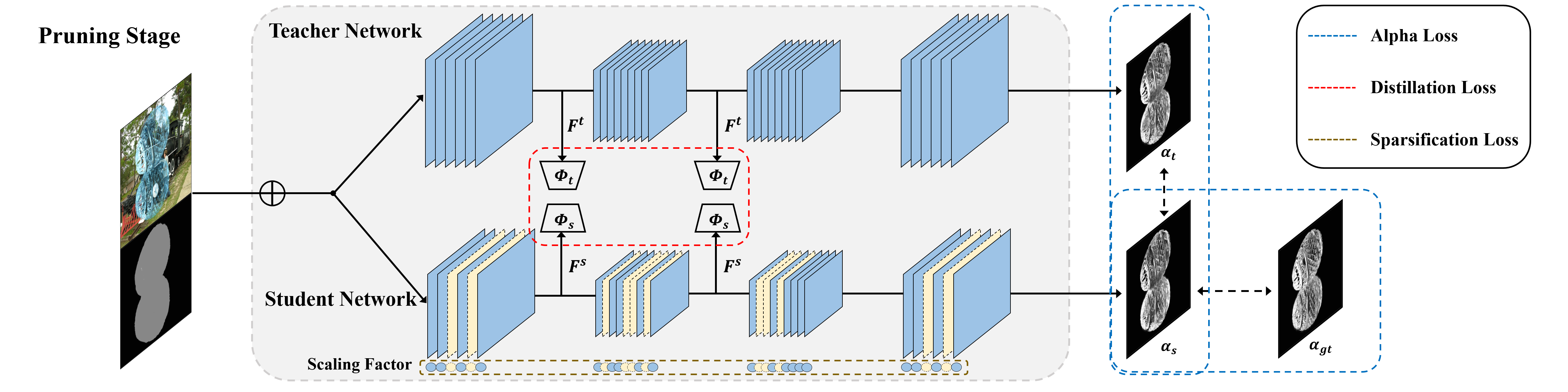} \\
    \includegraphics[width=0.94\linewidth]{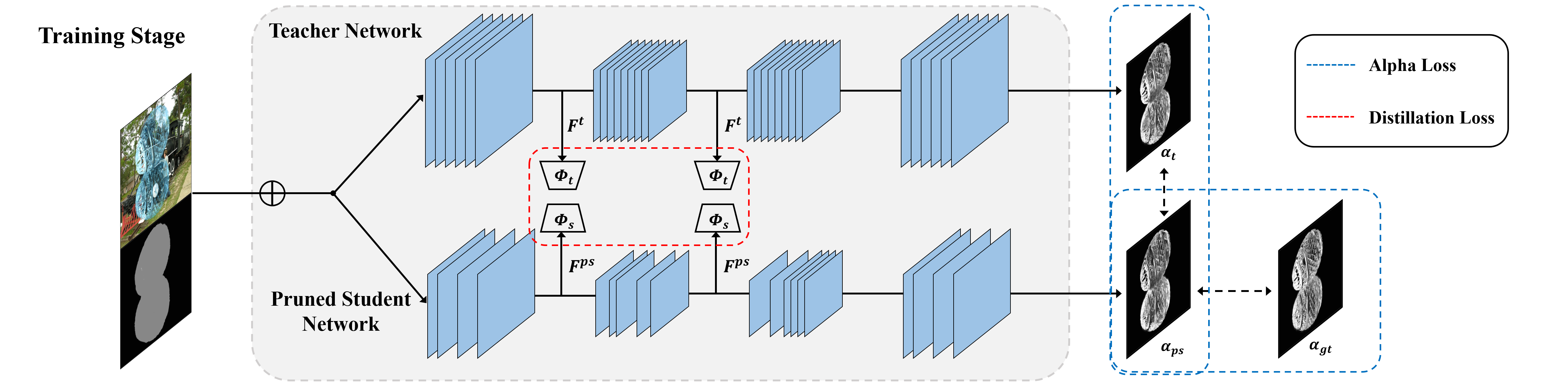} 
    \end{tabular}
    \caption{
    In the pruning stage, the student network is lightweighted using scaling factor sparsification loss and distillation loss with pre-trained teacher model. In the training stage, the same distillation loss is used to train the pruned student network. 
    %
    }
    \label{fig:overview}
\end{figure*}
\section{Proposed Approach}
In this section, we briefly describe the basics of KD and motivation of using KD for network pruning.
Then, we introduce the distillation-based channel pruning method for sparsifying alpha matting network, and explain a method for training pruned lightweight model with KD. 
We use the same distillation method for both pruning and training stages, even though it is also possible to utilize different methods.
Related experiments are provided by the ablation studies.
Overview of our distillation-based channel pruning and training is illustrated in~\figfref{fig:overview}.

\subsection{Knowledge Distillation}
Knowledge distillation (KD)~\cite{Hinton15distillation} is a technique supervising a small student model by a larger teacher model. 
The main purpose of KD is to transfer rich feature representations of the large model trained by the huge amount of data into the small model.
Therefore, it is very useful when there is a lack of training data or limited computational resources and memory of the devices.
%
%

Mathematically, feature maps of the teacher and student networks in $i$-th layer are denoted as $F^{t}_{i}\in\mathbb{R}^{C^{t}_{i}\times H^{t}_{i}\times W^{t}_{i}}$ and $F^{s}_{i}\in\mathbb{R}^{C^{s}_{i}\times H^{s}_{i}\times W^{s}_{i}}$, respectively.
Note that $\left \{C^{t}_{i}, C^{s}_{i} \right \}$ are the number of channels, $\left \{H^{t}_{i}, H^{s}_{i} \right \}$ and $\left \{W^{t}_{i}, W^{s}_{i} \right \}$ represent the spatial size.
Generally, the distillation loss for each layer is formulated as
\begin{equation}
    \mathcal{L}_{KD}(F^{t}_{i}, F^{s}_{i})  = \mathcal{L}_{F}(\Phi_{t}(F^{t}_{i}),\Phi_{s}(F^{s}_{i})),
    \label{eq:KD_loss}
\end{equation}
where, $\mathcal{L}_{F}(\cdot)$ is a similarity function, $\Phi_{t}(\cdot)$ and $\Phi_{s}(\cdot)$ are feature transform functions for the teacher and student networks. 
According to the purpose of distillation, appropriate  $\mathcal{L}_{F}(\cdot)$, $\Phi_{t}(\cdot)$ and $\Phi_{s}(\cdot)$ should be designed.
%
%

\subsection{Motivation}
Over the recent years, various KD methods have been introduced~\cite{Hinton15distillation,Yim17FSP,Tung19sim_distil,Heo19OFD,Ge19ZLL}, but most of them arbitrarily set the architecture of the student network.
%
%
%
Therefore, they do not ensure whether the student network is optimal for both distillation and the given tasks.
%
%
For example, the importance of each channel in the layers of a deep CNN model may be different, therefore, reducing the number of channels uniformly for all the layers is sub-optimal obviously.
%
%
We believe that it is also important for the alpha matting task to find the optimal student model.
%

To confirm this, we perform a preliminary experiment using GCA~\cite{Li20GCA} as a baseline matting model. 
%
%
First, we divide the encoder of GCA model into two groups: low-level layers (\texttt{conv1}-\texttt{conv3}) and high-level layers (\texttt{conv4}-\texttt{conv5}).
%
%
%
Then, we apply uniform 50\% channel pruning to low-level and high-level layers separately, and then obtain two different pruned networks.
The ratios of the removed channel parameters to the whole encoder are 12.75\% in low-level and 37.25\% in high-level layers, respectively.
In other words, more parameters are eliminated from the high-level layers rather than the low-level ones.
%
Using these two uniformly pruned GCA models, we verified the alpha matting performance on the Adobe-1k dataset~\cite{Xu17cvpr}.
\begin{table}[t]
    \addtolength{\tabcolsep}{1.2pt}
    \centering
    \caption{
    The first row: model with channels removed from low-level layers. The second row: model with channels removed from high-level layers.
    %
    }
    \begin{tabular}{cccccc}
    \toprule
    Methods & MSE     & SAD   & Grad  & Conn  & \#Param \\ \midrule
    Low-level pruned  & 0.012   & 45.22 & 24.85 & 39.77 & 22.63M  \\
    High-level pruned & \textbf{0.011}   & \textbf{40.33} & \textbf{20.34} & \textbf{35.48} & \textbf{8.83M}   \\
    \bottomrule
\end{tabular}
\label{table:motivation}
\end{table}
%
As reported in~\tabfref{table:motivation}, the model removing channels of high-level layers (the second row) prunes more channels than the model removing channels of low-level layers (the first row) but achieves better performance.
The number of network parameters is also much less.
As a result, it can be seen that the channels of low-level layers have more influence on the alpha matting performance.
Regarding that the high-level pruned network has achieved better performance, this result implies that the channel distributions of the original and optimal pruned models might be different significantly and the low-level layers are highly important in the alpha matting problem.
%

Therefore, in this paper, we propose a method to find a student network model that can well receive low-level knowledge from a large-capacity teacher network.
To this end, we introduce a distillation-aware pruning loss in the pruning stage to create an optimal lightweight network for alpha matting.
%
%
%
%
\subsection{Pruning with KD}
Inspired by~\cite{Liu2017slimming}, we adopt a channel pruning method based on the sparsification of scaling factors in batch normalization (BN) layers. 
BN layer is used in most deep CNN models for better generalization and fast convergence.
Formally, the BN layer is defined as follows:
\begin{equation}
    y = \gamma\frac{x-\mu}{\sqrt{\sigma^{2}+\epsilon}} + \beta,
    \label{eq:BN}
\end{equation}
where $x$ and $y$ are input and output of BN layer, and $\mu$ and $\sigma$ are mean and standard deviation of the input mini-batch features, and $\epsilon$ is a small constant. $\gamma$ and $\beta$ are the learnable scaling and shifting factors.
In~\cite{Liu2017slimming}, the scaling factor $\gamma$ in the BN layer is considered as the measure for the importance of each channel.
%
%
In other words, a channel with a very small $\gamma$ is regarded as the layer which does not contribute significantly to the final prediction.
%
%
%
Therefore, enforcing sparsification on the scaling factors eases the identification of prunable layers.

Similarly, our pruning method trains a target student network with sparsification loss and distillation loss, then remove channels with small scaling factors in BN layers.
We adopt the same alpha matting model for both teacher and student networks.
For the network pruning, only parameters of the student network are updated while those of the teacher network are fixed.
The final loss includes alpha prediction, channel sparsification, and distillation losses as follows:
%
%
\begin{eqnarray}
\mathcal{L}_{P}  =  \lambda_{1}\mathcal{L}_{\alpha}(\alpha_\mathrm{s},\alpha_\mathrm{gt}) + \lambda_{2}\mathcal{L}_{\alpha}(\alpha_\mathrm{s},\alpha_{\mathrm{t}}) + \lambda_{3}\sum_{\gamma \in\zeta} \left | \gamma \right | + \lambda_{4}\sum_{i\in\eta}{\mathcal{L}_{KD}(F^{t}_{i}, F^{s}_{i})},
\label{eq:pruning_loss}
\end{eqnarray}
where $\mathcal{L}_{\alpha}(\cdot)$ is the vanilla alpha prediction loss introduced in~\cite{Xu17cvpr}, and $\alpha_\mathrm{s}$, $\alpha_\mathrm{t}$, $\alpha_\mathrm{gt}$ are alpha matte prediction results from the student network and the teacher network, and ground truth, respectively. 
$\zeta$ is the set of scaling factors over all the BN layers and $\eta$ is the index set of layers utilized for distillation loss. 
%
%
$\lambda_{1}$, $\lambda_{2}$, $\lambda_{3}$, and $\lambda_{4}$ are balancing factors for each term.
Note that the distillation loss is used only for the encoder part.

In~\eqnref{eq:pruning_loss}, the gamma value $\gamma$ corresponding to the importance score can be estimated significantly differently depending on the distillation loss, which means that different pruned networks can be created.
Therefore, we adopt several recent KD methods to be utilized in the proposed channel pruning method as follows:
%
%
%
%
%

\begin{itemize}
\item \textbf{Neuron Selectivity Transfer (NST)}: Huang and Wang proposed NST~\cite{Huang2017NST} that aligns the distribution of spatial neuron activations between teacher and student networks. 
To this end, NST minimizes maximum mean discrepancy (MMD) distance between activations of teacher and student networks. 
Thus, $\Phi(\cdot)$ in~\eqnref{eq:KD_loss} is a certain function for kernel trick which projects samples into a higher dimensional feature space.
Also, $\mathcal{L}_{F}(\cdot)$ is distance (\ie, $L_{2}$ distance) between means of projected features of teacher and student networks.
\item \textbf{Overhaul of Feature Distillation (OFD)}: 
Heo~\etal~\cite{Heo19OFD} investigated various aspects of the existing feature distillation methods and suggested OFD that is a simple but effective distillation method.
In particular, $\Phi_{t}(\cdot)$ in~\eqnref{eq:KD_loss} is a margin ReLU function while $\Phi_{s}(\cdot)$ is a regressor consisting of a 1$\times$1 convolution layer.
Also, $\mathcal{L}_{F}(\cdot)$ in~\eqnref{eq:KD_loss} is a partial $L_{2}$ distance.
\item \textbf{Similarity-Preserving Knowledge Distillation (SPKD)}: 
The SPKD-based distillation method makes the pairwise similarity of the student network similar to that of the teacher network.
In~\cite{Tung19sim_distil}, batch similarity for the classification task is used while spatial and channel similarity for the regression task is utilized in~\cite{yoon2020lightweight,ko2022lowlight}.
Thus, $\Phi(\cdot)$ in~\eqnref{eq:KD_loss} is a function of making pairwise similarities and $\mathcal{L}_{F}(\cdot)$ is the $L_{2}$ distance.
\end{itemize}
%
%
%
%
%
After training with distillation loss as in~\eqnref{eq:pruning_loss}, we prune the target student network based on scaling factors of BN layers.
The smaller the scaling factor is, the less impact it has on the output of the layer, thus we remove the channels with a lower scaling factor than a threshold.
%
%
To eliminate $M$ channels, we adopt the $M$-th smallest scaling factor as the threshold.
%
At this point, thresholds of the encoder and decoder are obtained separately since distillation loss is only used in the encoder.
After pruning, we can get the compact lightweight alpha matting network, which is suitable to get fine details by KD.
\subsection{Training with KD}
By the aforementioned our distillation-based channel pruning, the architecture of a lightweight student network can be obtained. 
%
%
In \cite{liu2018rethinking}, the network structure itself is considered more important than the remaining parameters after pruning.
In other words, the fine-tuned model and the trained from scratch model achieve similar result, or even the trained from scratch model performs better.
Thus, we train the pruned network from scratch again by applying KD using the teacher network based on the loss function defined as follows:
%

\begin{eqnarray}
\mathcal{L}_{T} = w_{1}\mathcal{L}_{\alpha}(\alpha_\mathrm{ps},\alpha_\mathrm{gt}) + w_{2}\mathcal{L}_{\alpha}(\alpha_\mathrm{ps},\alpha_{\mathrm{t}}) + w_{3}\sum_{i\in\eta}{\mathcal{L}_{KD}(F^{t}_{i}, F^{ps}_{i})},
\label{eq:training_loss}
\end{eqnarray}
where $\alpha_\mathrm{ps}$ is a prediction of the pruned student network and $F^{ps}_{i}$ is feature maps in the $i$-th layer of the pruned student network.
$w_{1}$, $w_{2}$, and $w_{3}$ are balancing factors for each term in~\eqref{eq:training_loss}.
Unlike~\eqnref{eq:pruning_loss}, sparsification loss is not included, and the pruned student network is used.
We use the same distillation loss as the pruning step, but other distillation losses can be used. 

\begin{table*}[t]
    \centering
    \caption{
    Quantitative evaluation by GCA-50\% model on the benchmark.
    %
    }
    {\small
    \begin{tabular}{>{\centering}m{1.1cm}|>{\centering}m{1.2cm}>{\centering}m{1.2cm}|>{\centering}m{1cm}>{\centering}m{1cm}>{\centering}m{1cm}>{\centering}m{1cm}>{\centering}m{1.3cm}>{\centering\arraybackslash}m{1.3cm}}
    \toprule
    \multirow{2}{*}{Dataset}          & \multicolumn{2}{c|}{Methods}           & \multirow{2}{*}{MSE} & \multirow{2}{*}{SAD} & \multirow{2}{*}{Grad} & \multirow{2}{*}{Conn} & \multirow{2}{*}{\#Param} & \multirow{2}{*}{FLOPs} \\
                                      & KD                & Prune             &                      &                      &                       &                       &                          & \\
    \midrule
    \multirow{11}{*}{\rotatebox{90}{Adobe-1k}} & \multicolumn{2}{c|}{Unpruned(teacher)} & 0.009        & 35.28                & 16.92                 & 35.53                 & 25.27M  & 11.19G         \\
    \cmidrule{2-9}
                                      & -                 & UNI               & 0.017                & 52.61                & 34.27                 & 46.24                 & 6.35M     & \textbf{2.90G}        \\
                                      & -                 & NS                & 0.017                & 51.57                & 28.70                 & 45.66                 & \textbf{4.30M}     & 6.84G        \\
                                      & -                 & FBS               & 0.013                & 45.09                & 22.87                 & 39.39                 & 7.11M     & \textbf{2.90G}        \\
                                      & -                 & CAP               & 0.014                & 48.55                & 26.01                 & 42.76                 & 5.13M     & 6.50G        \\
    \cmidrule{2-9}
                                      & NST               & NS                & 0.020                & 58.10                & 36.78                 & 51.80                 & \textbf{4.30M}     & 6.84G        \\
                                      & NST               & Ours              & 0.017                & 53.47                & 30.59                 & 47.37                 & 5.80M     & 7.02G        \\
    \cmidrule{2-9}
                                      & OFD               & NS                & 0.014                & 46.96                & 24.16                 & 41.93                 & \textbf{4.30M}     & 6.84G        \\
                                      & OFD               & Ours              & 0.012                & 43.15                & 21.79                 & 37.66                 & 5.13M     & 6.48G        \\
    \cmidrule{2-9}
                                      & SPKD              & NS                & 0.012                & 42.69                & 21.88                 & 37.54                 & \textbf{4.30M}     & 6.84G        \\
                                      & SPKD              & Ours              & \textbf{0.011}       & \textbf{41.26}       & \textbf{21.42}        & \textbf{35.87}        & 4.66M     & 6.74G        \\
    \midrule
    \multirow{12}{*}{\rotatebox{90}{Disticntions-646}} & \multicolumn{2}{c|}{Unpruned(teacher)} & 0.025 & 27.60             & 15.82                 & 22.03                 & 25.27M   & 11.19G \\
    \cmidrule{2-9}
                                      & -                 & UNI               & 0.037                & 37.83                & 26.35                 & 28.36                 & 6.35M  & \textbf{2.90G} \\
                                      & -                 & NS                & 0.028                & 31.45                & 20.47                 & 25.69                 & \textbf{3.99M}  & 6.41G \\
                                      & -                 & FBS               & \textbf{0.024}       & 28.61                & \textbf{17.36}        & 23.52                 & 7.11M  & \textbf{2.90G} \\ 
                                      & -                 & CAP               & 0.026                & 29.23                & 18.59                 & 23.85                 & 5.30M  & 6.36G \\              
    \cmidrule{2-9}
                                      & NST               & NS                & 0.038                & 36.13                & 24.33                 & 28.20                 & \textbf{3.99M}  & 6.41G \\
                                      & NST               & Ours              & 0.035                & 36.58                & 23.69                 & 27.42                 & 4.93M  & 6.16G\\
    \cmidrule{2-9}
                                      & OFD               & NS                & 0.026                & 28.45                & 19.86                 & 23.22                 & \textbf{3.99M}  & 6.41G \\
                                      & OFD               & Ours              & \textbf{0.024}       & \textbf{27.12}       & \textbf{17.36}        & 22.03                 & 5.03M  & 6.22G\\
    \cmidrule{2-9}
                                      & SPKD              & NS                & \textbf{0.024}      & 27.88                 & 18.33                 & 22.46                 & \textbf{3.99M}  & 6.41G \\
                                      & SPKD              & Ours              & \textbf{0.024}      & 27.31                 & 17.81                 & \textbf{21.90}        & 4.24M  & 6.72G\\ 
    \bottomrule
    \end{tabular}%
    }
    \label{table:quantitative_GCA}
\end{table*}
\begin{figure*}[t]
\begin{center}
\def\arraystretch{0.4}
\begin{tabular}{@{}c@{}c@{}c@{}c@{}c@{}c@{}c}
\includegraphics[width=0.14\linewidth]{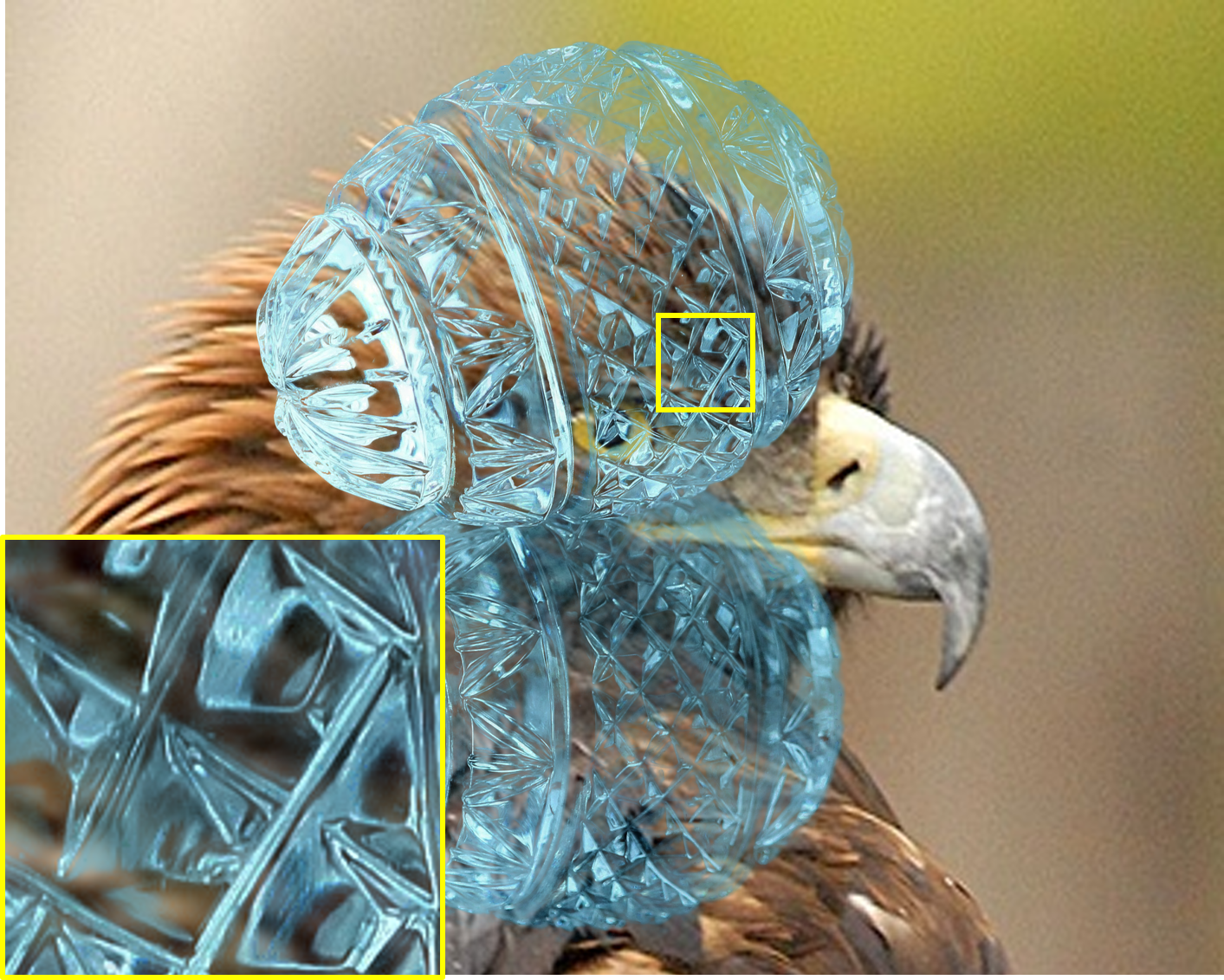} &
\includegraphics[width=0.14\linewidth]{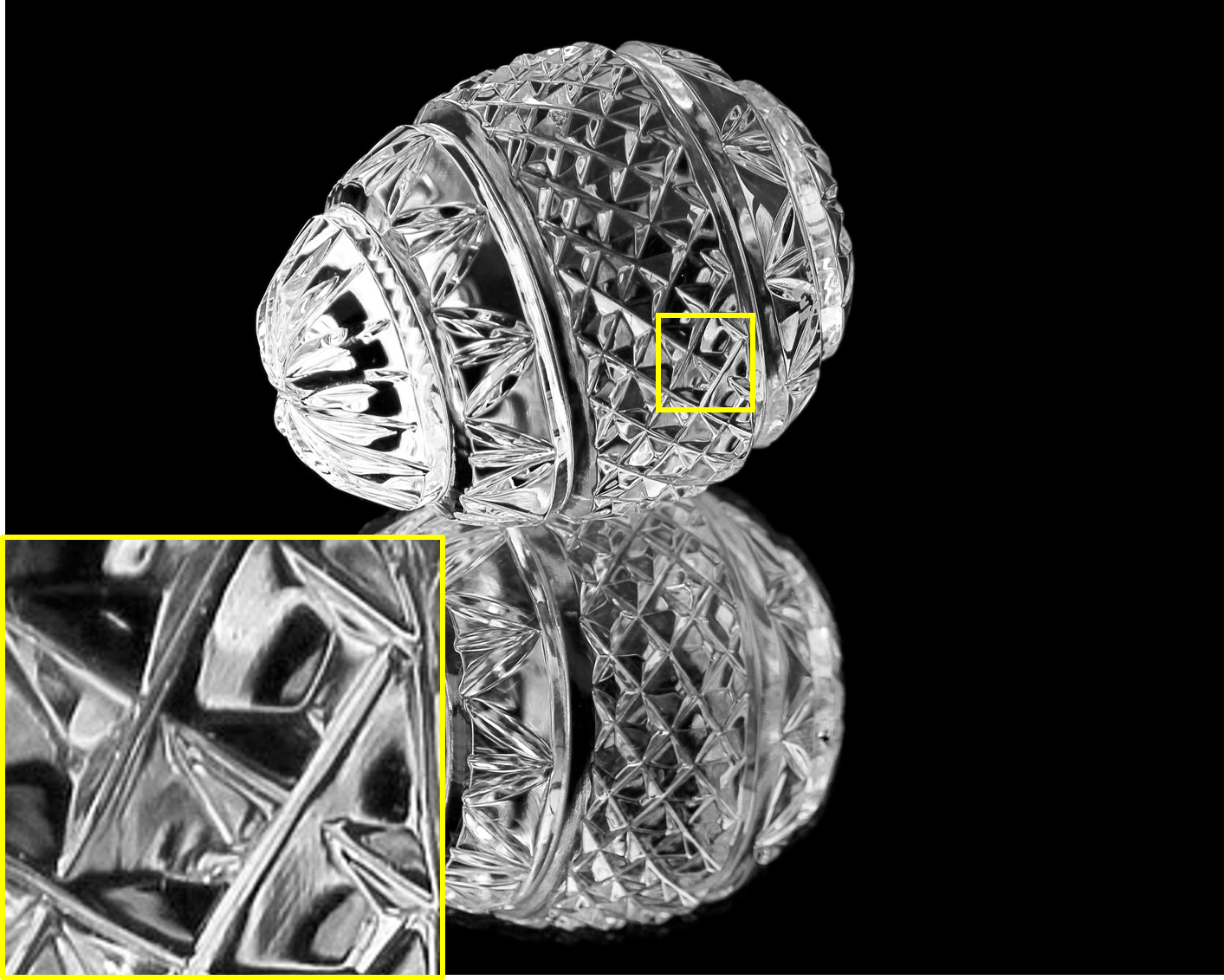} &
\includegraphics[width=0.14\linewidth]{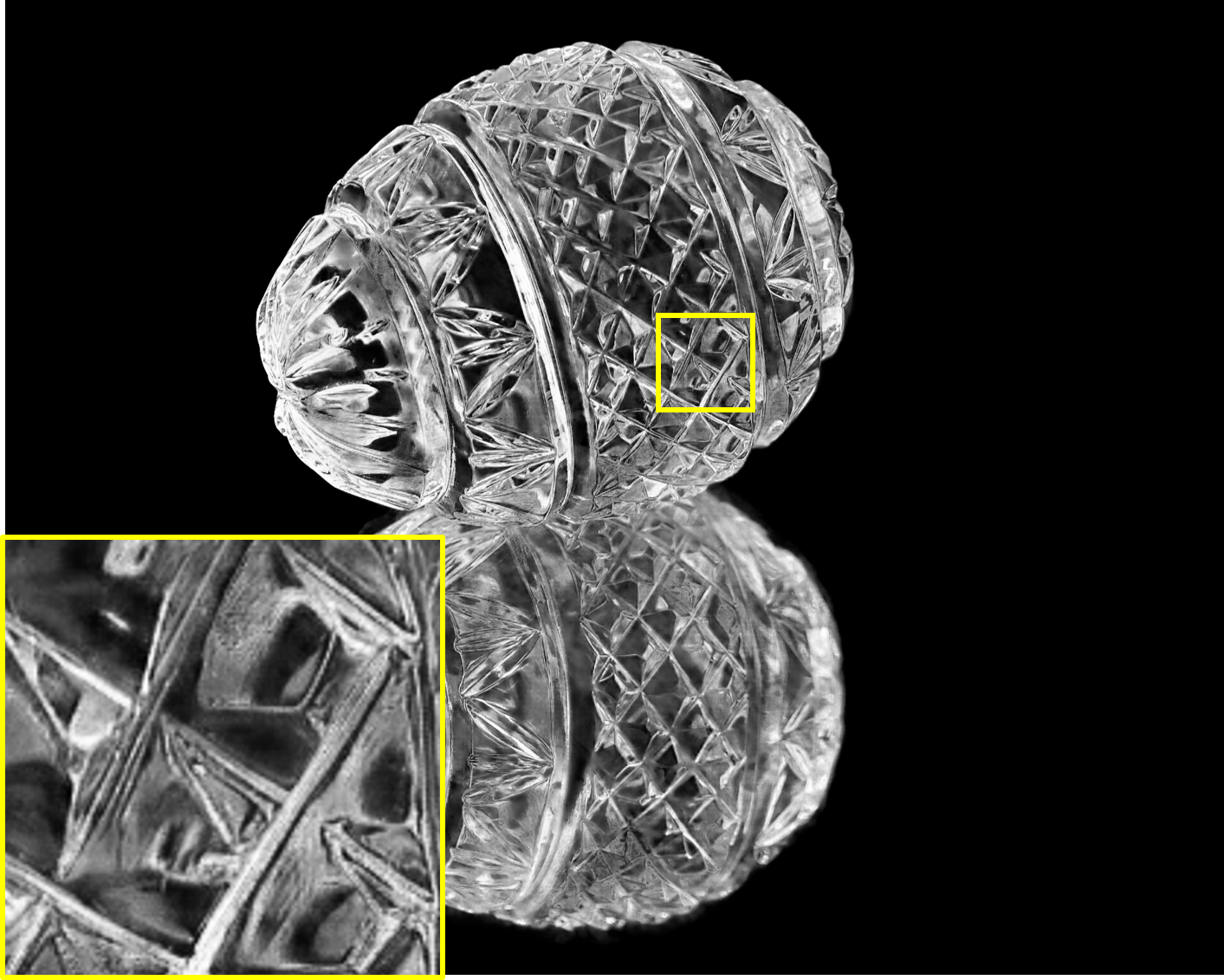} &
\includegraphics[width=0.14\linewidth]{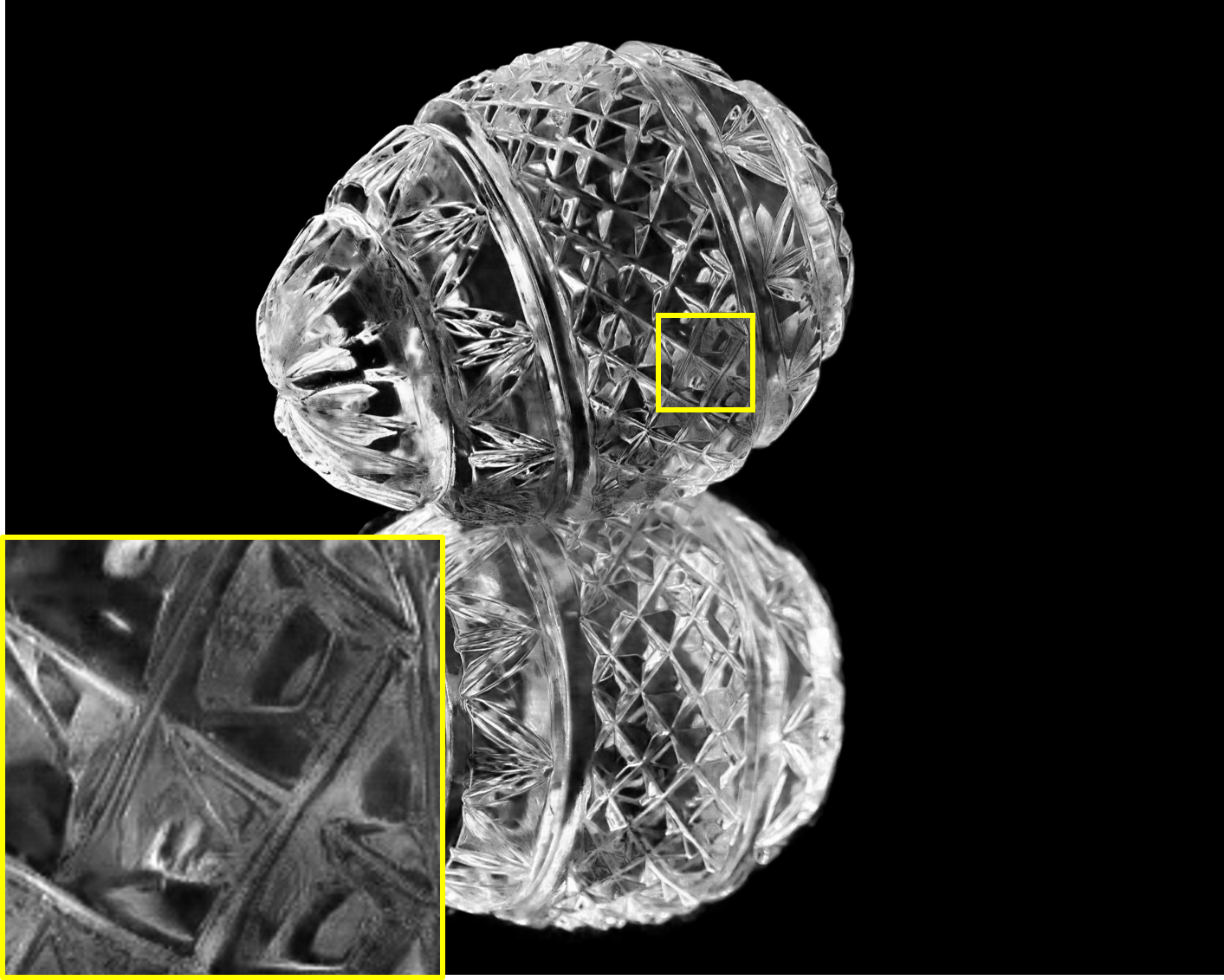} &
\includegraphics[width=0.14\linewidth]{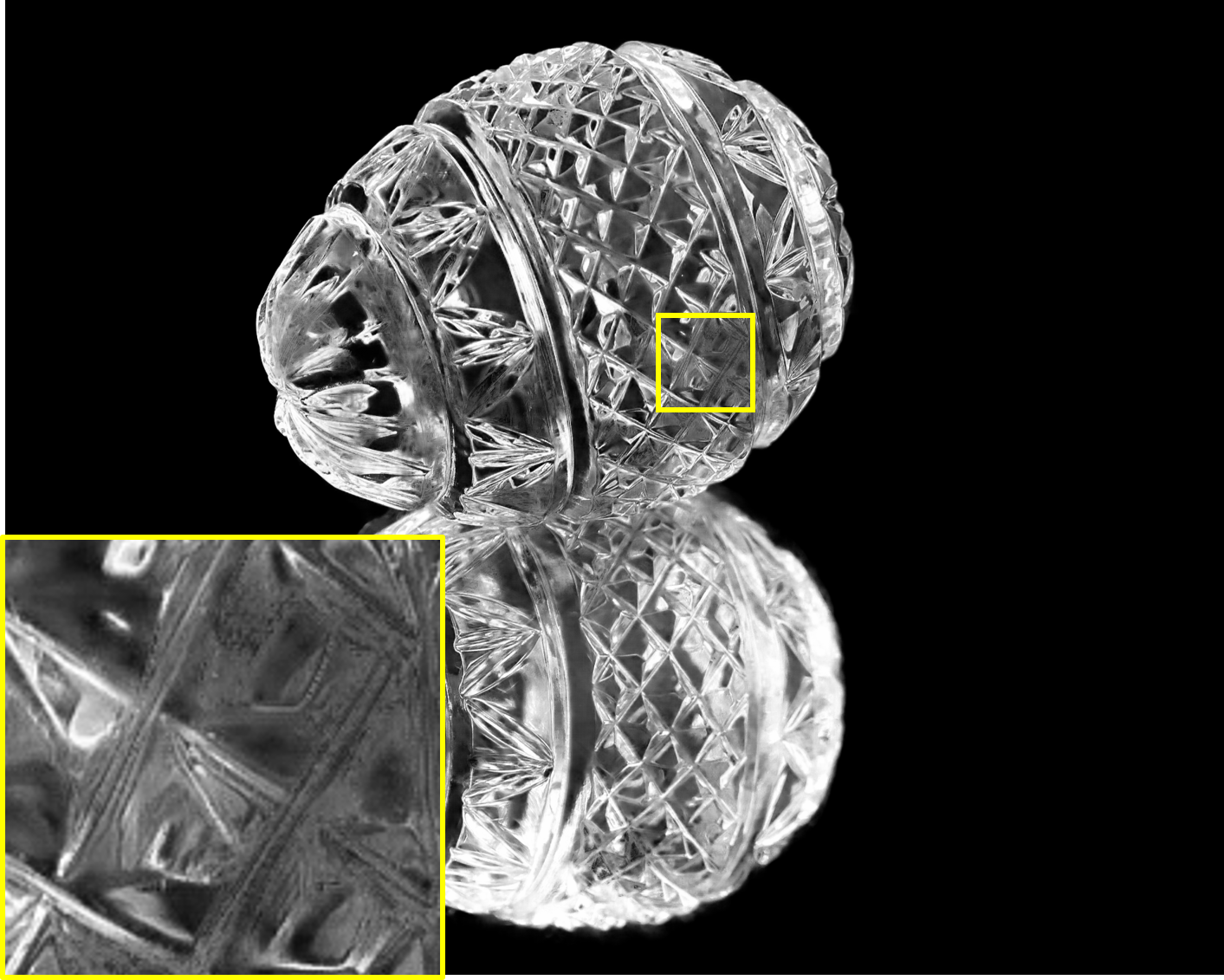} &
\includegraphics[width=0.14\linewidth]{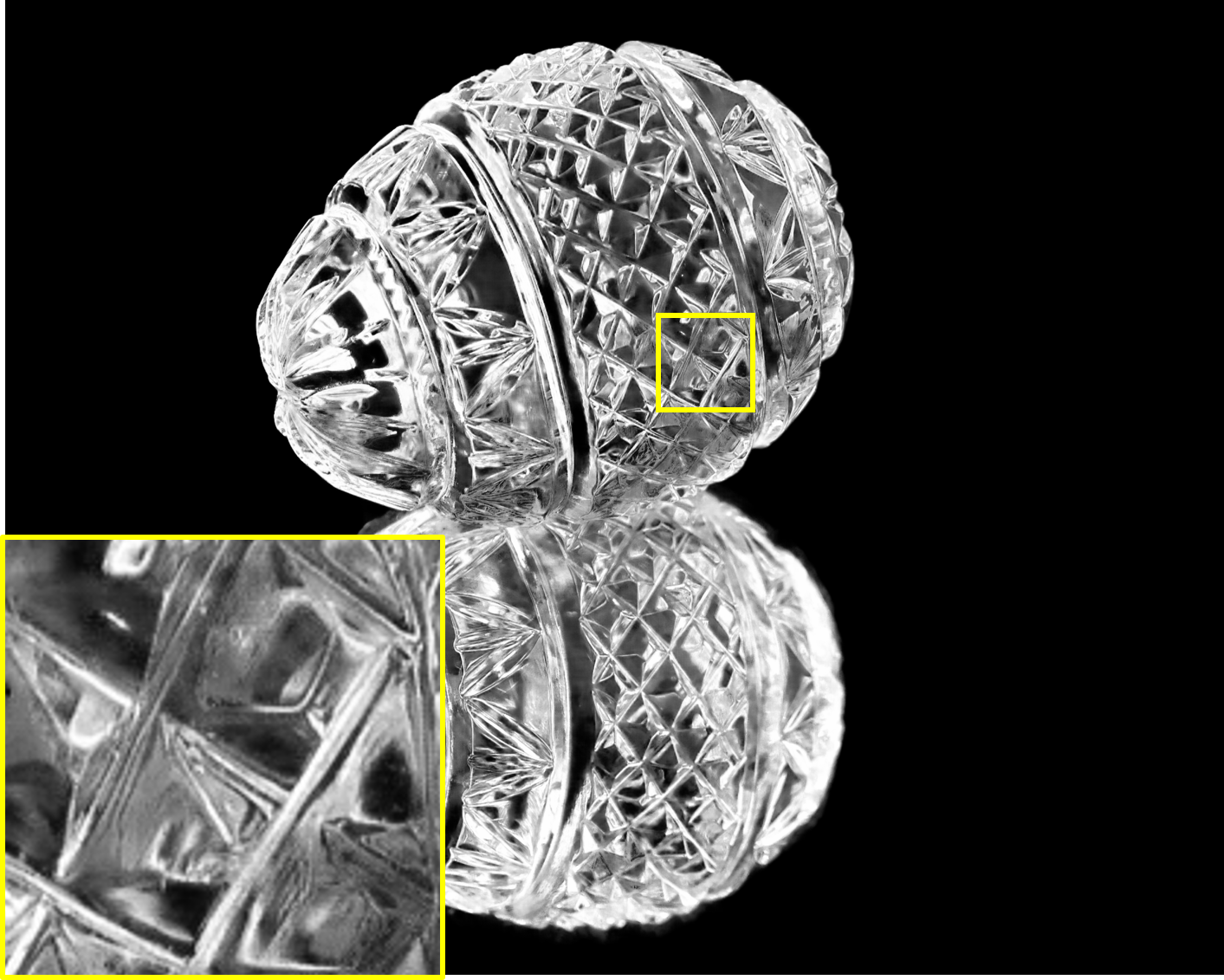} &
\includegraphics[width=0.14\linewidth]{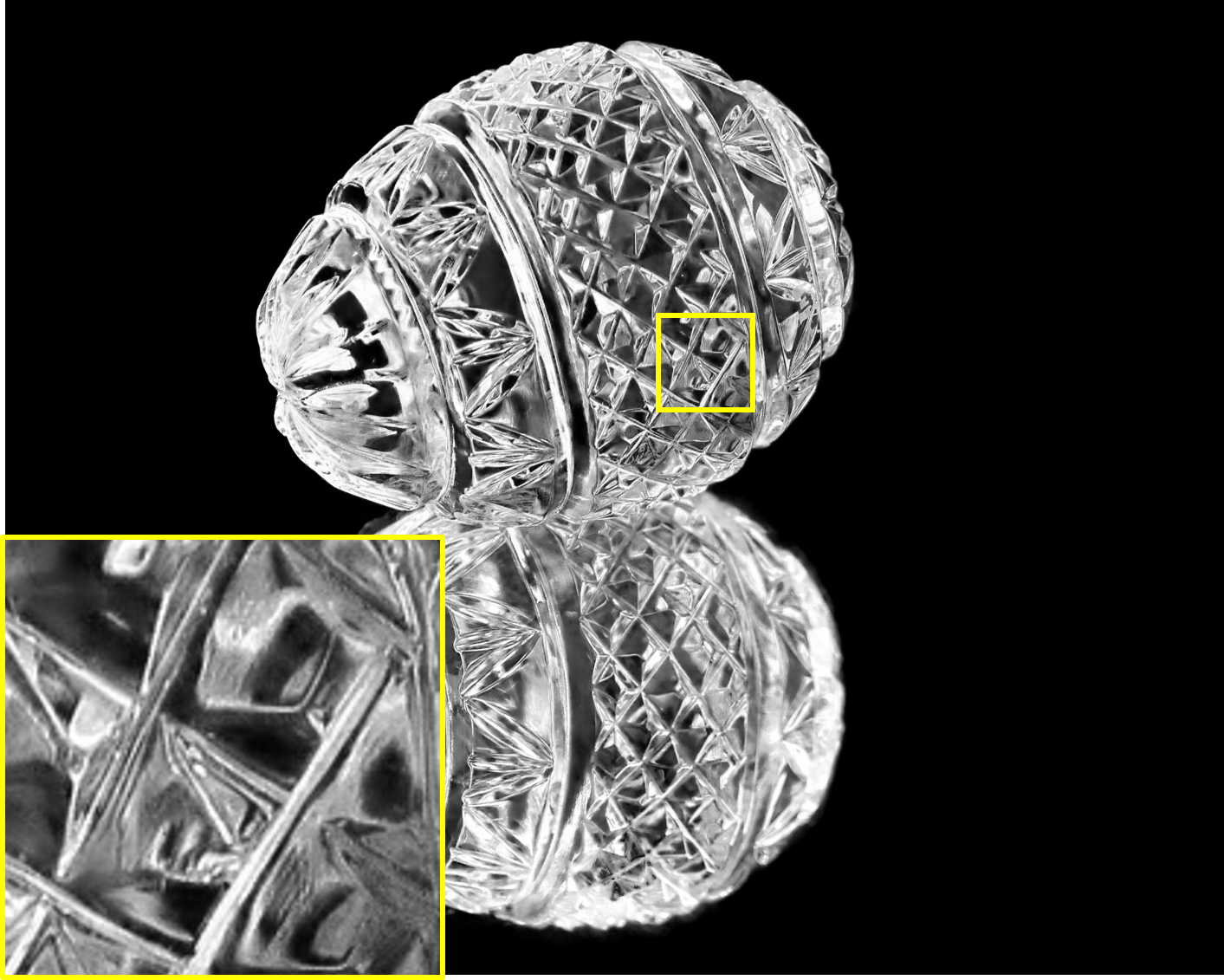}\\
\includegraphics[width=0.14\linewidth]{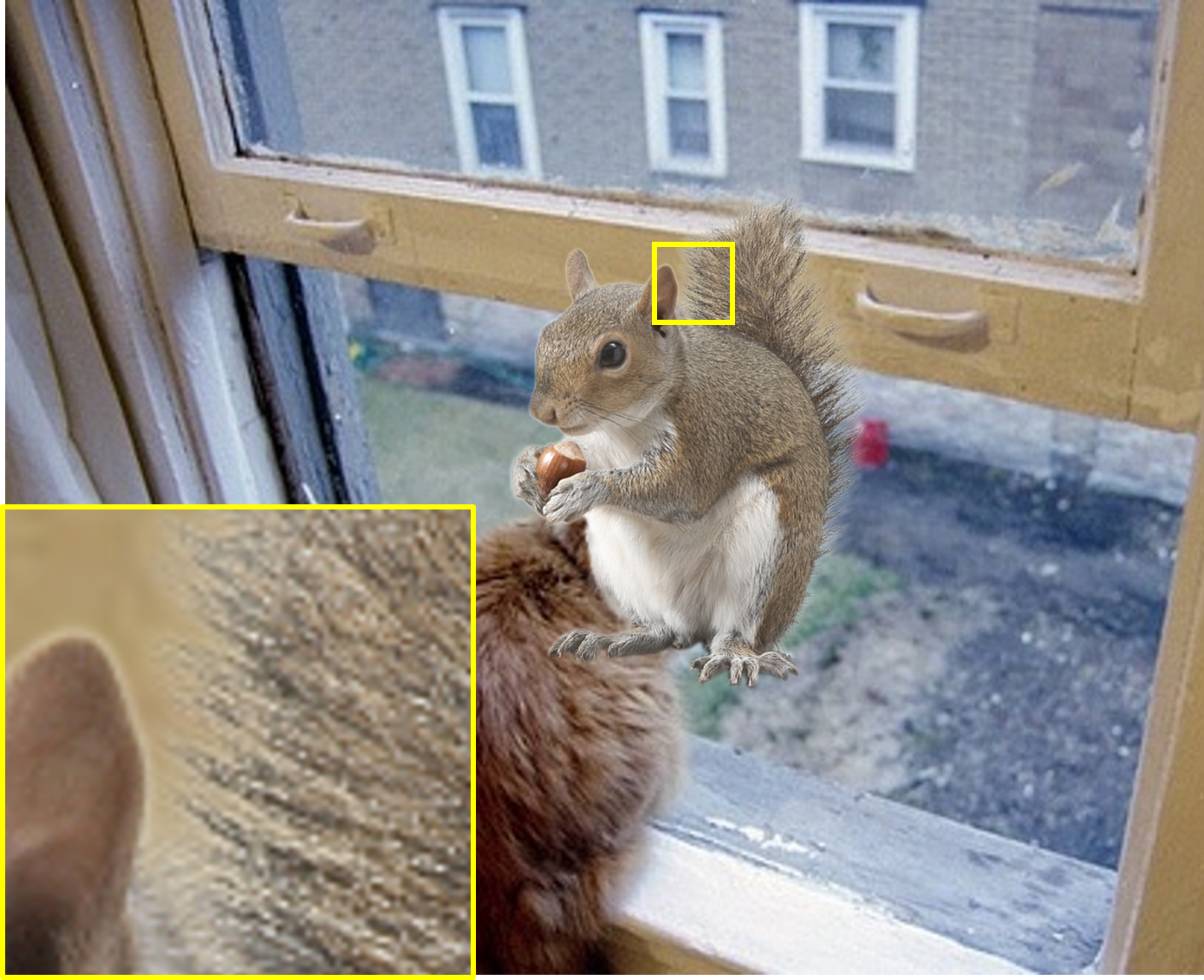} &
\includegraphics[width=0.14\linewidth]{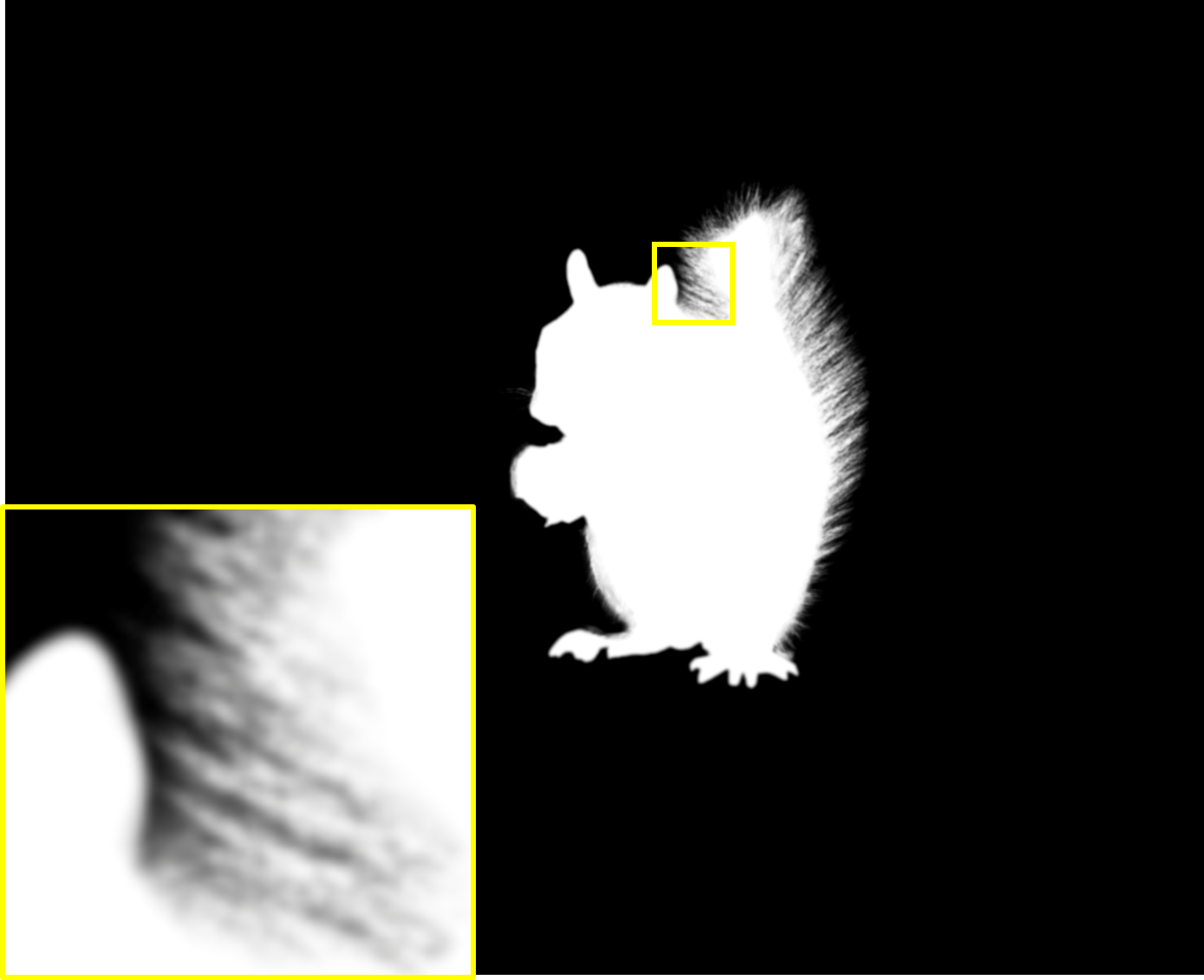} &
\includegraphics[width=0.14\linewidth]{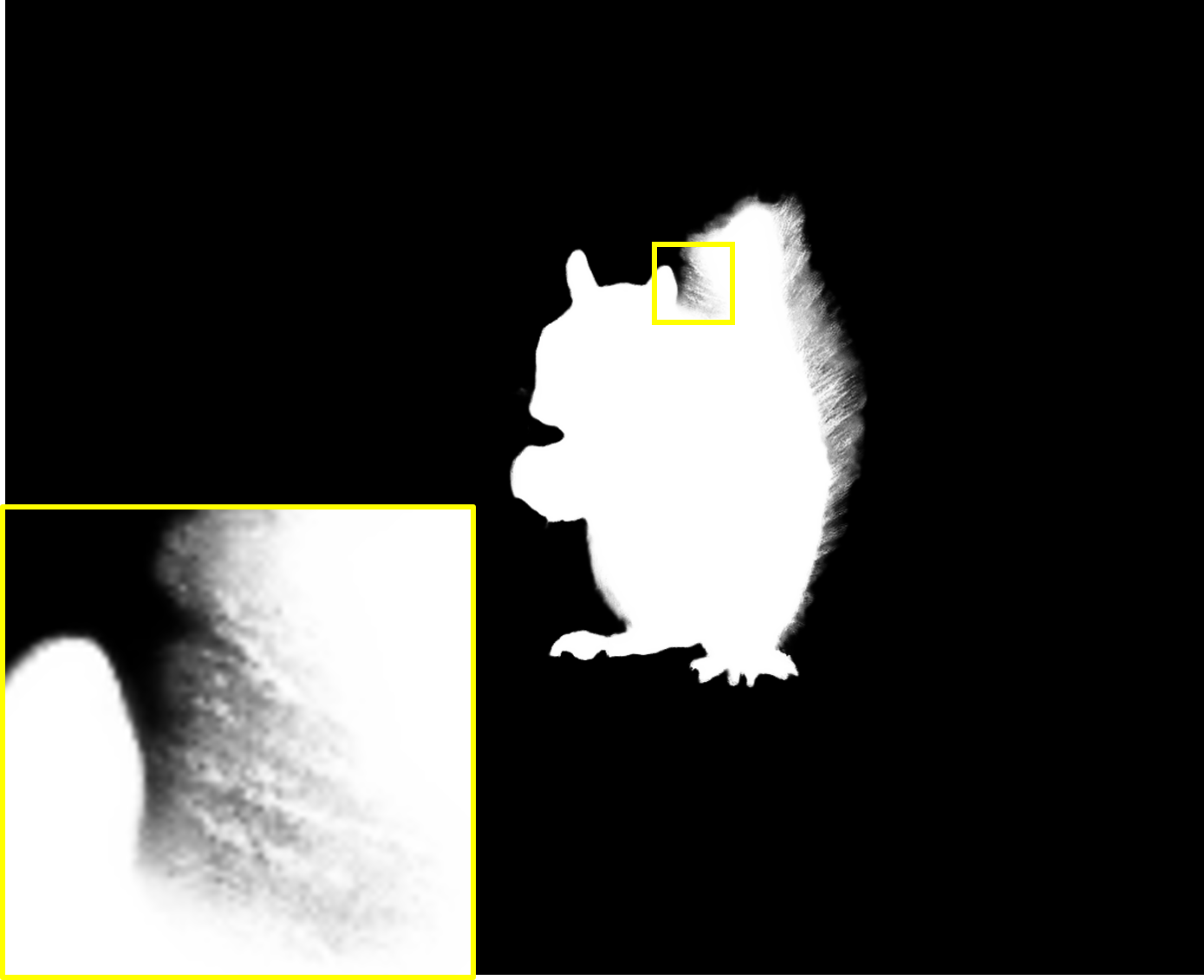} &
\includegraphics[width=0.14\linewidth]{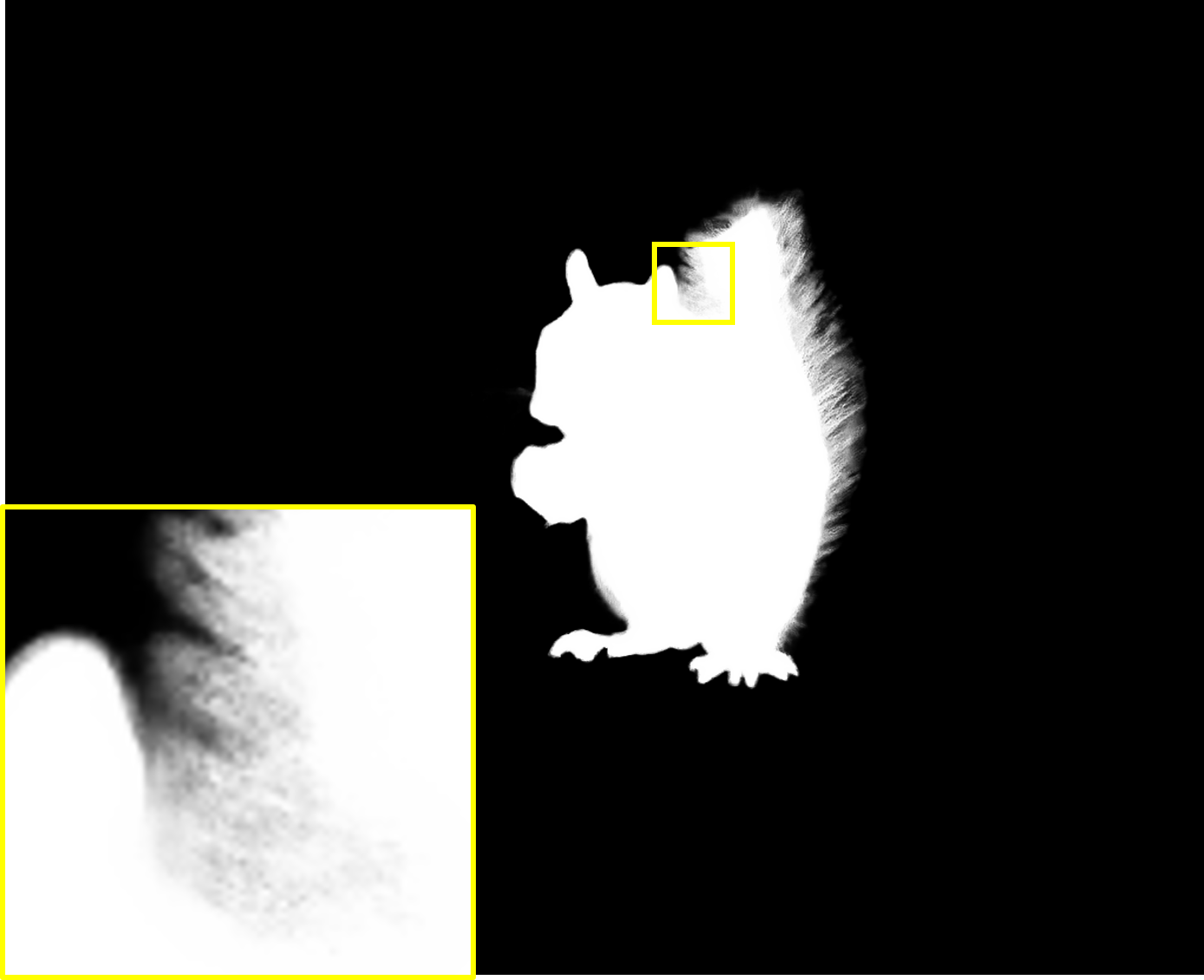} &
\includegraphics[width=0.14\linewidth]{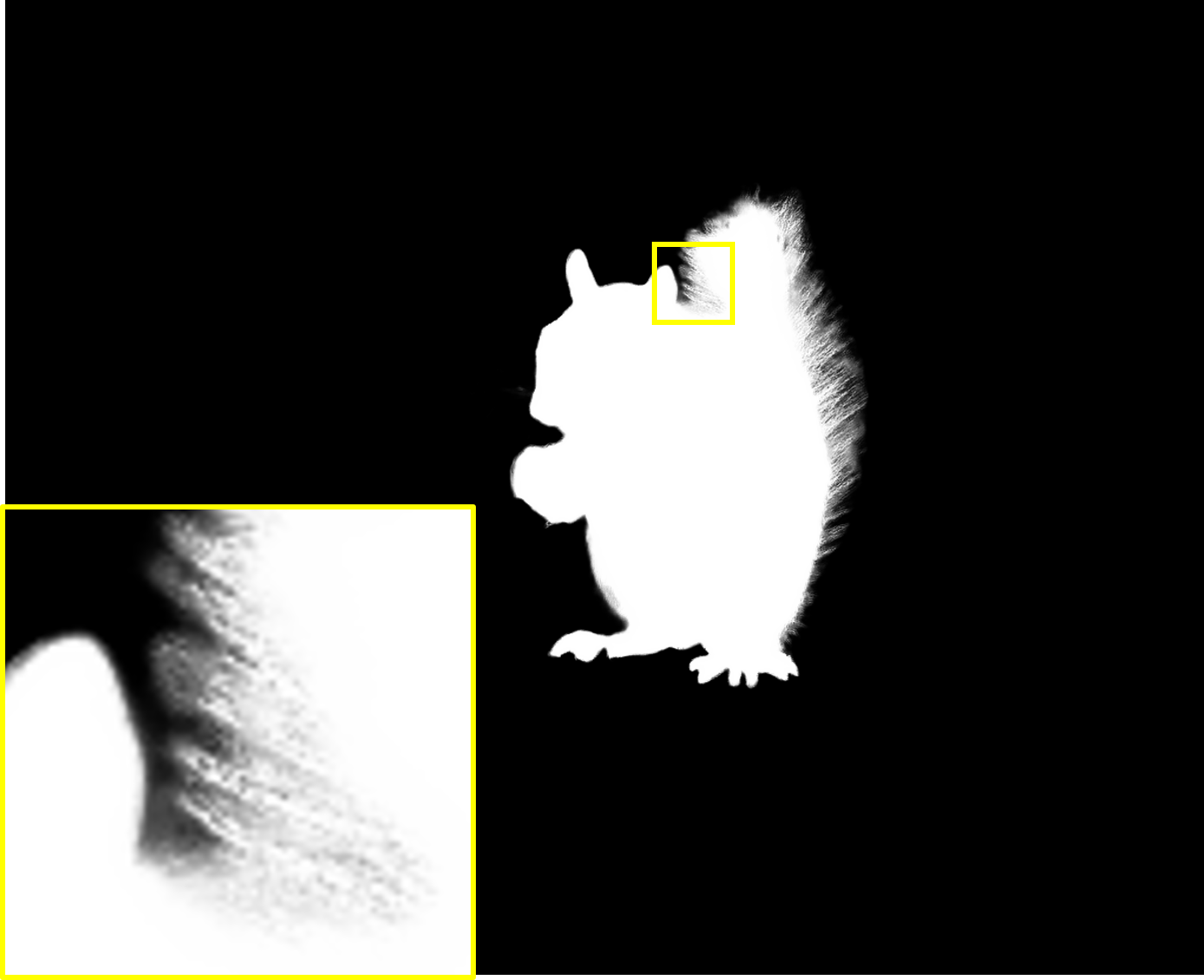} &
\includegraphics[width=0.14\linewidth]{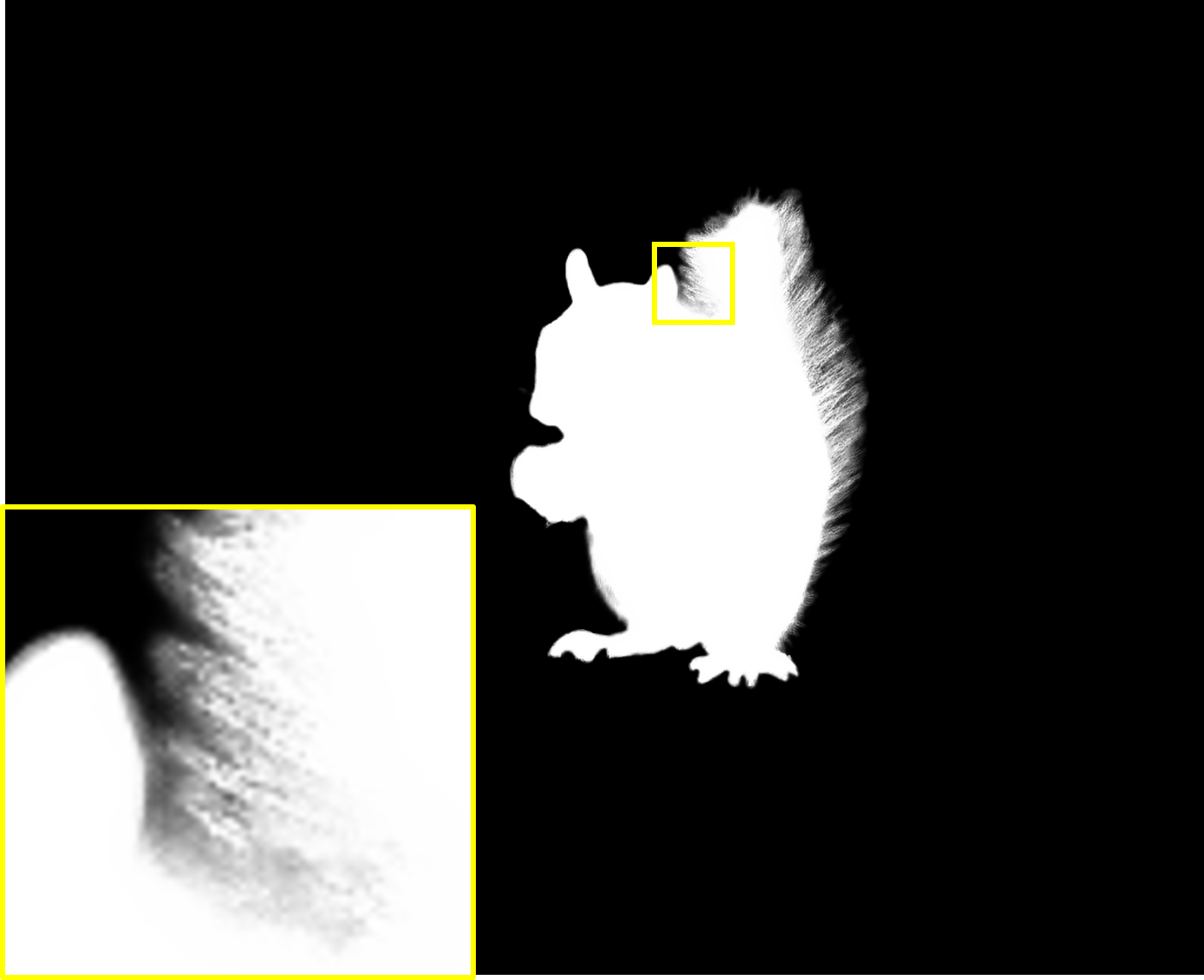} &
\includegraphics[width=0.14\linewidth]{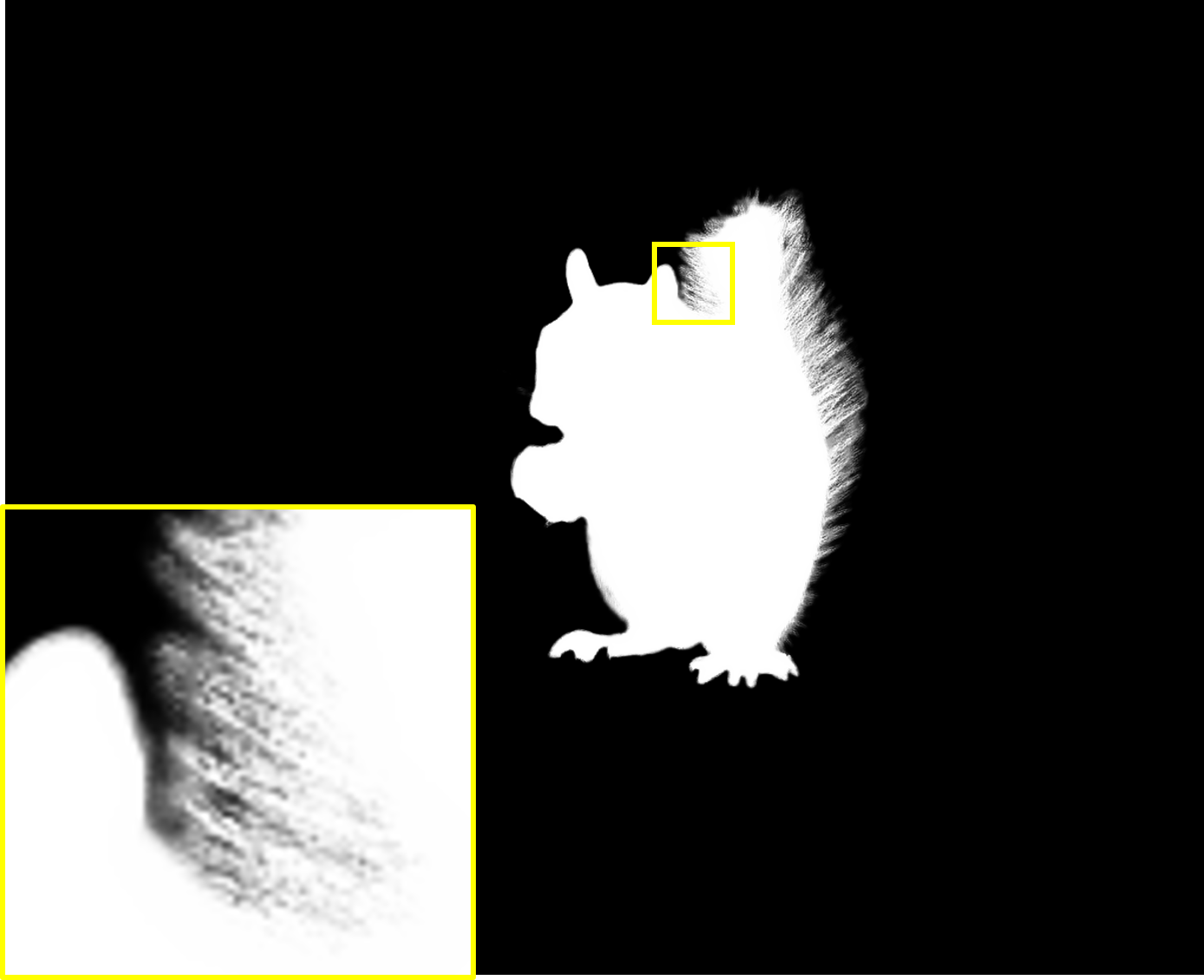} \\
{(a)} & {(b)} & {(c)} & {(d)} & {(e)} & {(f)} & {(g)}
\end{tabular}
\end{center}
\caption{Qualitative image matting results by GCA. (a) Input images. (b) Ground truths. (c) UNI. (d) NS~\cite{Liu2017slimming}. (e) CAP~\cite{He2021CAP}. (f) NS-SPKD. (g) Ours-SPKD.}
\label{fig:qualitative}
\end{figure*}

\begin{table}[t]
    \addtolength{\tabcolsep}{1.2pt}
    \centering
    \caption{Quantitative results by DIM-50\% model on the Adobe-1k.}
    \begin{tabular}{>{\centering}m{1.2cm}>{\centering}m{1.2cm}|>{\centering}m{1.1cm}>{\centering}m{1.1cm}>{\centering}m{1.1cm}>{\centering}m{1.1cm}>{\centering}m{1.3cm}>{\centering\arraybackslash}m{1.3cm}}
        \toprule
        \multicolumn{2}{c|}{Methods} & \multirow{2}{*}{MSE} & \multirow{2}{*}{SAD} & \multirow{2}{*}{Grad} & \multirow{2}{*}{Conn} & \multirow{2}{*}{\#Param} & \multirow{2}{*}{FLOPs} \\
         KD      & Prune            &                      &                      &                       &                       &                &  \\
        \midrule
        \multicolumn{2}{c|}{Unpruned(teacher)} & 0.021     & 65.37                & 33.20                 & 67.58                 & 25.58M       & 24.51G \\
        \midrule
                                   -       & UNI              & 0.049                & 114.02               & 78.89                 & 122.00                & 6.40M     & 6.19G     \\
                                   -       & NS               & 0.052                & 120.82               & 83.10                 & 129.70                & 6.32M     & 7.65G     \\
                                   -       & FBS              & 0.052                & 110.22               & 83.14                 & 118.60                & 6.92M     & 6.19G     \\
                                   -       & CAP              & 0.040                & 101.77               & 63.93                 & 108.61                & \textbf{4.08M}     & \textbf{4.26G}     \\
        \midrule
                                   NST     & NS               & 0.033                & 94.31                & 55.48                 & 100.67                & 6.32M     & 7.65G     \\
                                   NST     & Ours             & 0.032                & 89.83                & 49.04                 & 95.20                 & 5.98M     & 11.23G     \\
        \midrule
                                   OFD     & NS               & 0.029                & 76.74                & 42.45                 & 79.86                 & 6.32M     & 7.65G     \\
                                   OFD     & Ours             & 0.032                & 76.71                & 43.86                 & 80.61                 & 7.97M     & 17.61G     \\
        \midrule
                                   SPKD & NS                  & 0.029                & 76.73                & 42.70                 & 80.77                 & 6.32M     & 7.65G      \\
                                   SPKD & Ours                & \textbf{0.027}       & \textbf{73.67}       & \textbf{40.78}        & \textbf{76.58}        & 7.81M     & 12.53G     \\
        \bottomrule
    \end{tabular}%
    \label{table:quantitative_DIM}
\end{table}
\begin{table}[t]
    \addtolength{\tabcolsep}{0.8pt}
    \centering
    \caption{Quantitative results by IndexNet-25\% model on the Adobe-1k.}
    \begin{tabular}{>{\centering}m{1.2cm}>{\centering}m{1.2cm}|>{\centering}m{1.1cm}>{\centering}m{1.1cm}>{\centering}m{1.1cm}>{\centering}m{1.1cm}>{\centering}m{1.3cm}>{\centering\arraybackslash}m{1.3cm}}
        \toprule
        \multicolumn{2}{c|}{Methods} & \multirow{2}{*}{MSE} & \multirow{2}{*}{SAD} & \multirow{2}{*}{Grad} & \multirow{2}{*}{Conn} & \multirow{2}{*}{\#Param} & \multirow{2}{*}{FLOPs} \\
         KD      & Prune            &                      &                      &                       &                       &           &       \\
        \midrule
        \multicolumn{2}{c|}{Unpruned(teacher)} & 0.013     & 45.61                & 27.06                 & 43.79                 & 8.15M     & 5.64G   \\
        \midrule
                                   -       & UNI          & 0.027                & 65.21                & 42.56                 & 66.83                 & 3.49M      & \textbf{3.22G}    \\
                                   -       & NS               & 0.026                & 65.45                & 40.79                 & 67.11                 & 4.21M  & 5.34G         \\
                                   -       & FBS              & 0.020                & 56.43                & 34.41                 & 56.78                 & 4.05M  & \textbf{3.22G}         \\
                                   -       & CAP              & 0.018                & 53.32                & 35.17                 & 53.41                 & 4.07M  & 5.38G        \\
        \midrule
                                   NST     & NS               & 0.020                & 57.04                & 33.99                 & 57.49                 & 4.21M  & 5.34G         \\
                                   NST     & Ours             & 0.019                & 54.21                & 28.14                 & 53.57                 & \textbf{4.02M}  & 4.95G        \\
        \midrule
                                   OFD     & NS               & 0.017                & 52.13                & 29.99                 & 51.76                 & 4.21M  & 5.34G        \\
                                   OFD     & Ours             & 0.015                & 47.24                & \textbf{24.42}        & 46.26                 & 4.41M  & 5.13G        \\
        \midrule
                                   SPKD & NS            & 0.016                & 50.35                & 27.21                 & 49.86                 & 4.21M      & 5.34G          \\
                                   SPKD & Ours         & \textbf{0.014}        & \textbf{47.06}         & 25.98                 & \textbf{45.77}                 & 5.09M     & 5.21G           \\
    \bottomrule
    \end{tabular}%
    \label{table:quantitative_IndexNet}
\end{table}
\begin{table}[t]
    \centering
    \addtolength{\tabcolsep}{0.8pt}
    \caption{Ablation study on various combinations of methods for  distillation and pruning method. All evaluations are conducted by GCA-50\% model on the Adobe-1k.}
    \label{table:quantitative_mismatch}
    \begin{tabular}{>{\centering}m{1.2cm}>{\centering}m{1.4cm}|>{\centering}m{1.25cm}>{\centering}m{1.25cm}>{\centering}m{1.25cm}>{\centering\arraybackslash}m{1.25cm}}
    \toprule
    \multicolumn{2}{c|}{Methods}    & \multirow{2}{*}{MSE} & \multirow{2}{*}{SAD} & \multirow{2}{*}{Grad} & \multirow{2}{*}{Conn} \\
    KD                    & Prune &                      &                      &                       &                       \\
    \midrule
    \multirow{3}{*}{NST}  & +NST  & 0.017                & 53.47                & 30.59                 & 47.37                 \\
                          & +OFD  & 0.018                & 53.70                & 30.85                 & 46.78                 \\
                          & +SPKD & 0.019                & 56.85                & 35.94                 & 49.30                 \\
    \midrule
    \multirow{3}{*}{OFD}  & +NST  & 0.012                & 44.09                & 21.91                 & 39.24                 \\
                          & +OFD  & 0.012                        & 43.15                & 21.79                 & 37.66                 \\
                          & +SPKD & 0.013                & 44.14                & 23.51                 & 39.82                 \\
    \midrule
    \multirow{3}{*}{SPKD} & +NST  & 0.013                & 43.98                & 23.65                 & 38.82                 \\
                          & +OFD  & \textbf{0.011}                & 42.07                & 21.51                 & 36.58                 \\
                          & +SPKD & \textbf{0.011}                & \textbf{41.26}                & \textbf{21.42}                 & \textbf{35.87}                 \\
    \bottomrule
    \end{tabular}
    \label{table:unpaired}
\end{table}

\begin{table}[t]
    \addtolength{\tabcolsep}{0.8pt}
    \centering
    \caption{Results according to various pruning ratios. All evaluations are conducted by GCA-model with SPKD on the Adobe-1k.}
    \begin{tabular}{>{\centering}m{1.7cm}|>{\centering}m{1.2cm}>{\centering}m{1.2cm}>{\centering}m{1.2cm}>{\centering}m{1.2cm}>{\centering\arraybackslash}m{1.4cm}}
    \toprule
    Methods   & MSE     & SAD   & Grad  & Conn  & \#Param \\ \midrule
    UNI-30\%  & 0.012   & 43.46 & 23.78 & 38.61 & 12.00M  \\
    UNI-50\%  & 0.015   & 48.06 & 30.28 & 42.81 & 6.35M  \\
    UNI-70\%  & 0.019   & 58.26 & 33.53 & 50.77 & 2.50M  \\ \midrule
    Ours-30\% & \textbf{0.010}   & \textbf{39.38} & \textbf{19.45} & \textbf{34.80} & 10.37M \\
    Ours-50\% & 0.011   & 41.26 & 21.42 & 35.87 & 4.66M  \\
    Ours-70\% & 0.014   & 47.36 & 25.87 & 42.02 & \textbf{1.80M}   \\
    \bottomrule
\end{tabular}
\label{table:ratio}
\end{table}
%
\begin{table}[t]
\centering
    \addtolength{\tabcolsep}{0.8pt}
    \caption{Results of training our pruned model from scratch without  distillation.}
    \begin{tabular}{>{\centering}m{1.7cm}|>{\centering}m{1.2cm}>{\centering}m{1.2cm}>{\centering}m{1.2cm}>{\centering\arraybackslash}m{1.2cm}}
    \toprule
     Prune      & MSE     & SAD   & Grad  & Conn    \\ \midrule
     +NST       & \textbf{0.017}   & \textbf{54.02} & \textbf{31.50} & \textbf{47.45}   \\
     +OFD       & \textbf{0.017}   & 56.44 & 33.10 & 49.39   \\
     +SPKD      & \textbf{0.017}   & 56.88 & 32.62 & 50.09   \\
    \bottomrule
\end{tabular}
\label{table:prune_only}
\end{table}
%
\begin{table}[t]
    \centering
    \caption{Comparisons of Running Time (RT) per image on DIM model.}
    \begin{tabular}{>{\centering}m{2.0cm}|>{\centering}m{1.1cm}>{\centering}m{1.1cm}>{\centering}m{1.3cm}>{\centering}m{1.3cm}>{\centering\arraybackslash}m{1.2cm}}
        \toprule
        Method                  & MSE   & SAD    & \#Param & FLOPs & RT(ms)  \\ \midrule
        Unpruned                & 0.021 & 65.37  & 25.58M   & 24.51G & 11.33   \\ \midrule
        UNI                     & 0.049 & 114.02 & 6.40M    & 6.19G  & 3.80   \\
        NS                      & 0.052 & 110.22 & 6.32M    & 7.65G  & 3.87   \\ 
        CAP                     & 0.040 & 101.77 & \textbf{4.08M}    & \textbf{4.26G}  & \textbf{3.50}   \\ \midrule
        Ours(NST)               & 0.032 & 89.83  & 5.98M    & 11.23G & 5.53   \\
        Ours(OFD)               & 0.032 & 76.71  & 7.97M    & 17.61G & 5.67   \\
        Ours(SPKD)              & \textbf{0.027} & \textbf{73.67}  & 7.81M    & 12.53G & 5.64   \\ \bottomrule
        \end{tabular}
    \label{table:Running time}
\end{table}
%
%
\section{Experimental Results}
In this section, we evaluate the proposed distillation-based channel pruning method both quantitatively and qualitatively. 
We validate our method on various teacher models including GCA~\cite{Li20GCA}, DIM~\cite{Xu17cvpr}, and IndexNet~\cite{Hao19indexnet}, and also provide various ablation studies. 
Finally, we show that the proposed algorithm can be utilized for the other task such as semantic segmentation.
%
%
%
\subsection{Implementation Details}
%
In most experiments, we adopt GCA matting as a baseline alpha matting network.
In order to evaluate our distillation-based channel pruning method, we use two public benchmark datasets: Adobe-1k~\cite{Xu17cvpr} and Distinctions-646~\cite{Yu20AGHSA}.
Since Distinctions-646 test set does not provide official trimaps, we generate trimaps from ground truth alpha matte using dilation with kernel size 10.
The evaluation metrics for all quantitative experiments are mean squared error (MSE), sum of absolute difference (SAD), gradient error (Grad), connectivity (Conn), the number of network parameters (\#Param) and floating-point-opertions (FLOPs). 
%
%
We use activations of the last four layers in the encoder for computing distillation loss as in~\cite{yoon2020lightweight} for a fair comparison.

\subsection{Quantitative Comparisons}
%
We quantitatively verify our distillation-based channel pruning and training methods on Adobe-1k and Distinctions-646 datasets.
We adopt the aforementioned NST~\cite{Huang2017NST}, OFD~\cite{Heo19OFD}, and SPKD~\cite{yoon2020lightweight} as KD methods for both pruning and training stages.
For comparison, uniform channel pruning (UNI), network slimming (NS)~\cite{Liu2017slimming}, feature boosting and suppression (FBS)~\cite{gao2018dynamic}, and context-aware pruning (CAP)~\cite{He2021CAP} are chosen.
%
As reported in~\tabfref{table:quantitative_GCA}, the number of parameters in all student networks and FLOPs are much smaller than that of a teacher network (about 16-25\% parameters and 60\% FLOPs).
Although our pruned model sometimes has more parameters or more FLOPs than other pruned models (UNI, NS, FBS and CAP), the alpha matting performance is far superior to their performances.
Also, our distillation-based channel pruning method achieves better performance than NS regardless of distillation types.
Note that we utilize the same KD method in the training step for both our pruning method and NS.
Usually, performance is slightly higher when SPKD is used than OFD. 
However, when NST is used, the performance is lower than the existing pruning that does not include KD in the training step.
It indicates that the type of distillation loss is also an important factor for both the pruning and training.

To verify the generality of our method, the same experiments are performed using DIM and IndexNet as backbone models instead of GCA matting.
Similar to the case of GCA matting, the best performance is achieved with SPKD as reported in~\tabfref{table:quantitative_DIM} and~\tabfref{table:quantitative_IndexNet}. 
A different point from the case of GCA matting is that comparable performance was achieved even when using NST.
Note that the original IndexNet is already a lightweight model because it is based on MobileNetv2~\cite{Sandler_2018_CVPR}, but it can be even lighter by applying our channel pruning.
%

%
%
%
%
%
%
%
%

\subsection{Qualitative Comparisons}
\figfref{fig:qualitative} shows the qualitative performance of our method. 
We compare our results obtained using SPKD with results from existing pruning algorithms.
%
%
Examples contain various object structures: short hair, overlapped color distribution ($squirrel$) and transparency ($glass$). 
As expected, the results of the existing pruning methods are over-smoothed as shown in the ($glass$) example of \figref{fig:qualitative}-(d,e). In this example, UNI produces a better result than NS and CAP.  
%
%
%
Overall, results using distillation loss in the pruning step~(\figref{fig:qualitative}-(f)) show stable and visually pleasing predictions.
Moreover, our final results using the SPKD in both pruning and training step~(\figref{fig:qualitative}-(g)) provide the best predictions with fine details preserved.

\subsection{Ablation Studies}
%

\noindent{\textbf{Different Distillation for Pruning and Training.}}
Since it is possible to utilize different distillation losses for pruning and training stages, it is meaningful to explore whether it is better to use different distillation losses in pruning and training steps or to use the same distillation loss. %
To this end, we performed experiments on all combinations of NST, OFD, and SPKD in the pruning and training phases.
As reported in~\tabfref{table:unpaired}, we can achieve better performance when the same distillation loss is used in the pruning and training phases.
Even more, in the student model pruned with NST, it is better to use NST in the training stage than OFD and SPKD, which are more advanced distillation techniques.
%
These results are reasonable because the student network architecture obtained by a specific KD method will have a high chance to be more effective for the same distillation method than the other ones.

\noindent{\textbf{Pruning Ratio.}}
We analyze our distillation-based pruning according to pruning ratios.
We compare the results of the model in which the number of channels is reduced by 30$\%$, 50$\%$, and 70$\%$ using the our method, and the model uniformly reduced in the same proportion.
For all cases, we use the same SPKD as distillation loss for training step.
As in~\tabfref{table:ratio}, the pruned models whose channels are reduced by 70$\%$ and 50$\%$ using our method achieve better performance and fewer network parameters than the model uniformly pruned by 50$\%$.
%
%
%
%
%
%

\noindent{\textbf{Training from Scratch without KD.}}
To analyze the effect of paired distillation loss for both the pruning and training stages, we train our pruned model from scratch without KD. 
As reported in~\tabfref{table:prune_only}, our pruned model achieves slightly worse performance than models pruned by UNI, NS, and CAP when trained without KD in the training phase.
Therefore, we conclude that our distillation-based channel pruning is more beneficial when it is combined with the proper distillation method during the training.
%

\noindent{\textbf{Running Time}}
We measure the running time of the each pruned model using Adobe-1k dataset.
As reported in~\tabref{table:Running time}, the performance (MSE, SAD) of our method with SPKD is quite close to those of unpruned teacher model while it runs twice faster than the teacher model.
The existing methods (UNI, NS, CAP) are faster than our method, but the performance (MSE, SAD) is very poor.

\begin{figure}[t]
\begin{center}
\def\arraystretch{0.4}
\begin{tabular}{@{}c@{}c@{}c@{}c@{}c@{}c}
\includegraphics[width=0.16\linewidth]{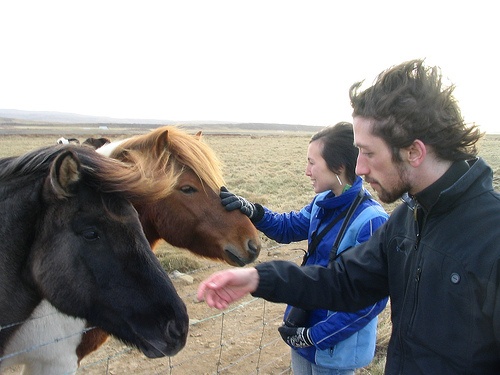} &
\includegraphics[width=0.16\linewidth]{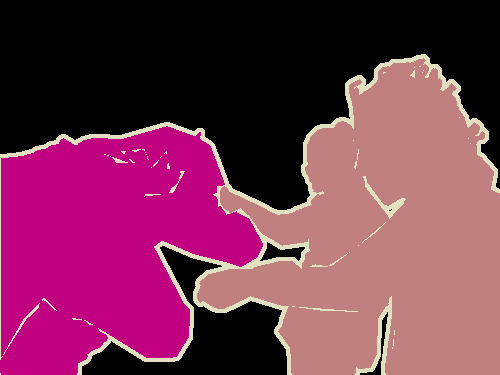} &
\includegraphics[width=0.16\linewidth]{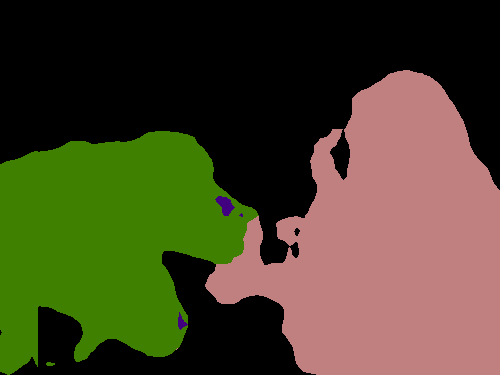} &
\includegraphics[width=0.16\linewidth]{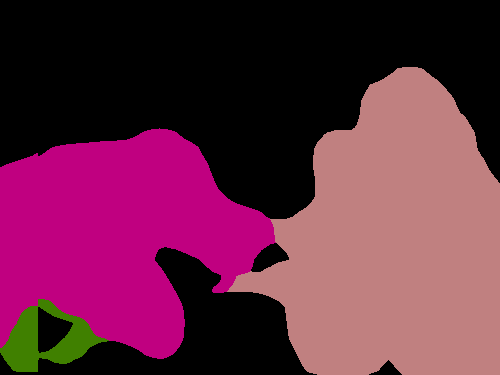} &
\includegraphics[width=0.16\linewidth]{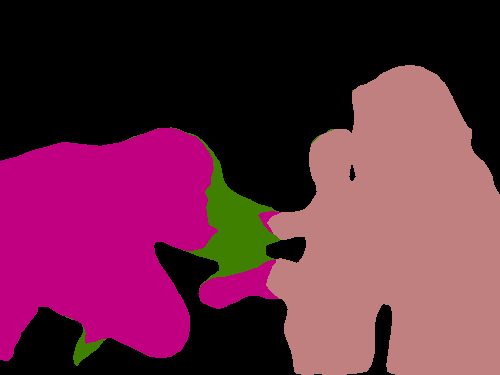} &
\includegraphics[width=0.16\linewidth]{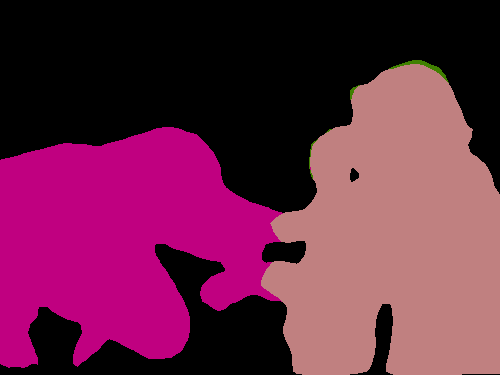} \\
\includegraphics[width=0.16\linewidth]{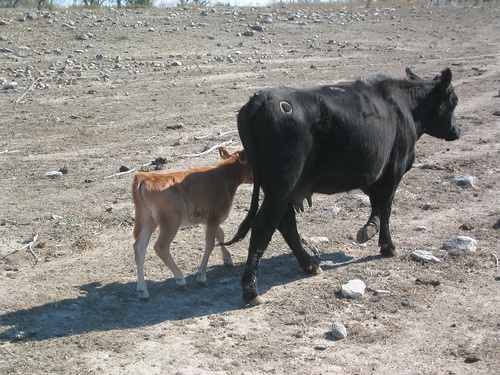} &
\includegraphics[width=0.16\linewidth]{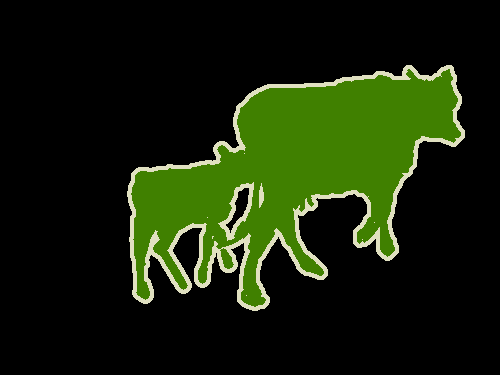} &
\includegraphics[width=0.16\linewidth]{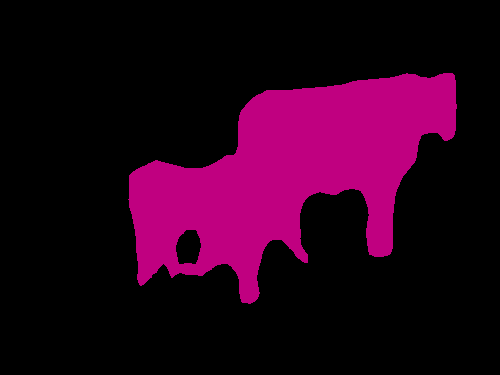} &
\includegraphics[width=0.16\linewidth]{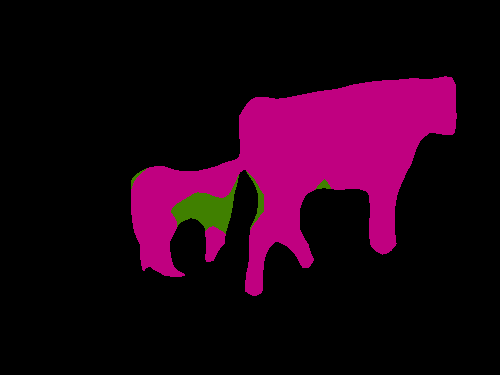} &
\includegraphics[width=0.16\linewidth]{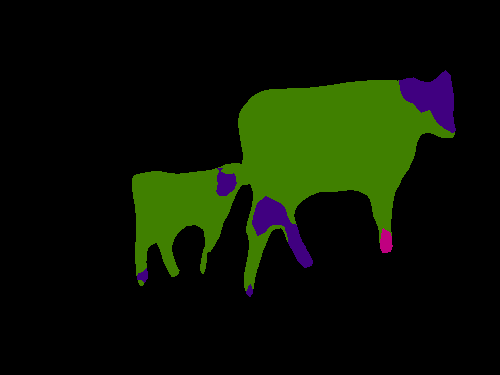} &
\includegraphics[width=0.16\linewidth]{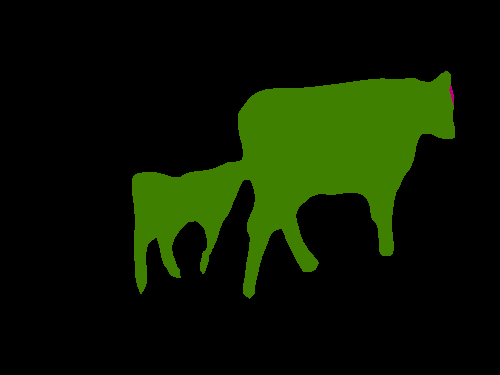} \\
{(a)} & {(b)} & {(c)} & {(d)} & {(e)} & {(f)}
\end{tabular}
\end{center}
\caption{Example of semantic segmentation results on PASCAL VOC2012 validation set. (a) Input images. (b) Ground truths. (c) NS. (d) CAP. (e) NS-OFD. (f) Ours-OFD.}
\label{fig:pascal}
\end{figure}
\begin{table}[t]
    \addtolength{\tabcolsep}{0.8pt}
    \centering
    \caption{Semantic segmentation results by 50$\%$ pruned PSPNet-50.}
    \begin{tabular}{>{\centering}m{0.9cm}>{\centering}m{0.9cm}|>{\centering}m{0.9cm}>{\centering}m{0.9cm}>{\centering}m{1.2cm}||>{\centering}m{0.9cm}>{\centering}m{0.9cm}|>{\centering}m{0.9cm}>{\centering}m{0.9cm}>{\centering\arraybackslash}m{1.2cm}}
    \toprule
    \multicolumn{2}{c|}{Methods}  & \multirow{2}{*}{mIoU} &\multirow{2}{*}{Acc.} & \multirow{2}{*}{\#Param} & \multicolumn{2}{c|}{Methods}  & \multirow{2}{*}{mIoU} &\multirow{2}{*}{Acc.} & \multirow{2}{*}{\#Param}\\
    KD            & Prune        &                  &           &           & KD            & Prune        &                  &          &                    \\ \midrule
    \multicolumn{2}{c|}{\multirow{2}{*}{\shortstack{Unpruned\\(teacher)}}} & \multirow{2}{*}{78.02}    & \multirow{2}{*}{95.13}    & \multirow{2}{*}{49.1M}             & NST           & NS           & 60.67            & 90.80    & 26.5M                   \\ 
    \multicolumn{2}{c|}{}           &                  &           &           & NST           & Ours         & 60.81            & 90.69    & 22.0M                  \\ \hline
    -             & UNI          & 51.40            & 88.05     & \textbf{12.3M}     & OFD           & NS           & 62.45            & 90.69    & 26.5M                  \\
    -             & NS           & 41.78            & 85.24     & 26.5M     & OFD           & Ours         & \textbf{63.48}   & \textbf{90.85} & 21.7M            \\ \cline{6-10}
    -             & FBS          & 53.62            & 88.75     & 19.7M     & SPKD          & NS           & 58.26            & 89.55    & 26.5M                  \\
    -             & CAP          & 53.61            & 88.58     & 22.2M     & SPKD          & Ours         & 58.20            & 89.18    & 21.9M                  \\ 
    \bottomrule
    \end{tabular}%
\label{tab:quantitative_segmentation}
\end{table}

\noindent{\textbf{Application on Semantic Segmentation.}}
Our distillation-based channel pruning technique is applicable not only to alpha matting but also to other tasks. 
Therefore, in this subsection, we verify whether the proposed method is effective for semantic segmentation.
For experiments, we adopt PSPNet-50~\cite{Zhao_2017_CVPR} as a baseline model and test our method on Pascal VOC 2012 validation set~\cite{Everingham2014ThePV}.
We utilize the mean Intersection over Union (mIoU), and pixel accuracy (Acc.) as evaluation metrics.
As reported in~\tabfref{tab:quantitative_segmentation}, the proposed distillation-based channel pruning method achieves superior performance compared to the other existing channel pruning methods. 
Note that the performance of the pruned model by the NS is similar to our method when KD is applied, but the pruned model by our method has much fewer parameters.
Also, as shown in~\figfref{fig:pascal}, our channel pruning method produces a visually more pleasing result compared to the other channel pruning techniques.
%

\section{Conclusion}
We have proposed a distillation-based channel pruning method for lightening a deep image matting network. 
In the pruning step, we train a student network that has the same architecture with a teacher network using the distillation-based sparsification loss.
Then, we remove channels that have low scaling factor of BN layer.
Finally, we train the pruned student network using the same distillation loss utilized in the pruning step. 
Experimental results demonstrate that our distillation-based channel pruning method successfully reduces the number of parameters.
The lightweight network obtained by the proposed method achieves significantly better performance than other lightweight networks with similar capacity.
%
%
We analyze the proposed channel pruning technique through extensive ablation studies.

\section*{Acknowledgement}
This work was partly supported by Institute of Information \& communications Technology Planning \& Evaluation (IITP) grant funded by the Korea government(MSIT) (No.RS-2022-00155857, Artificial Intelligence Convergence Innovation Human Resources Development (Chungnam National University)) and the National Research Foundation of Korea (NRF) grant funded by the Korea government. (MSIT, No.2021R1A4A1032580 and No.2022R1C1C1009334).

\clearpage
\bibliographystyle{splncs04}
\bibliography{main}

\end{document}